\documentclass[11pt]{article}
\usepackage[preprint]{acl}
\usepackage{times}
\usepackage{latexsym}
\usepackage[T1]{fontenc}
\usepackage[utf8]{inputenc}
\usepackage{microtype}
\usepackage{graphicx}
\usepackage{booktabs}
\usepackage{amsmath}
\usepackage{amssymb}
\usepackage{array}

\title{Many Biases, One Circuit: Post-Training Tunes a Pretrained \\ Cue-Routing Mechanism}

\author{
\textbf{Prakhar Gupta\textsuperscript{1,2}}\thanks{Email: \texttt{prakharg@umich.edu}.} \quad
\textbf{Terry Jingchen Zhang\textsuperscript{2,3,4}} \quad
\textbf{Zhijing Jin\textsuperscript{2,4,5}} \\[0.15em]
\textsuperscript{1}University of Michigan \quad
\textsuperscript{2}Jinesis Lab, University of Toronto \& Vector Institute\\
\textsuperscript{3}University of Oxford \quad
\textsuperscript{4}EuroSafeAI\\
\textsuperscript{5}Max Planck Institute for Intelligent Systems, T\"{u}bingen, Germany
}

\begin{document}
\raggedbottom

\maketitle

\begin{abstract}
Language models often abandon a correct answer when the prompt contains a cue, a hint toward a wrong option such as ``I think the answer is B.'' Sycophancy and related biases show this failure. They are often studied separately and blamed on alignment. We test this view causally, tracing the attention heads behind these biases through the stages of two post-training pipelines. In multiple-choice tests on most models, four cue types rely largely on the same few heads, a shared circuit already present in pretrained models. Most base models follow cues in plain text. Fine-tuning a base model on about 200 generic assistant examples without cues activates these heads in chat format, so cue-following can be a side effect of assistant tuning. The final reinforcement learning (RL) stage strengthens the effect of cues in almost every comparison. In Tulu-3-8B and OLMo-2-7B, we swap the attention patterns of six middle layers between the RL checkpoint and its predecessor. For suggested-answer cues, this shifts each checkpoint's cue effect most of the way to the other's level. Post-training thus seems to change when and how strongly models use this circuit. Mitigation for these multiple-choice biases could target this circuit.
\end{abstract}

\section{Introduction}
\label{sec:intro}

Language models often give up a correct answer when the prompt contains a cue. A \emph{cue} is text added to the prompt that points to one answer option without giving real evidence for it. For example, a model that answers a question correctly may switch to option B after the user adds the sentence ``I think the answer is B.'' We call such a switch to the cued wrong option \emph{caving}. Caving is a simple and measurable form of sycophancy \citep{sharma2024sycophancy,perez2022discovering}, and it also affects the reasoning given for the answer \citep{turpin2023language}. Other cues also cause caving, such as a fake earlier answer attributed to the model, an irrelevant fact about a wrong option, or a few-shot example in which the same question has a wrong label \citep{chua2024bct}. These are usually treated as separate biases. In multiple-choice tests, we find that they rely mostly on the same few attention heads.

\begin{figure*}[t]
\centering
\includegraphics[width=\textwidth]{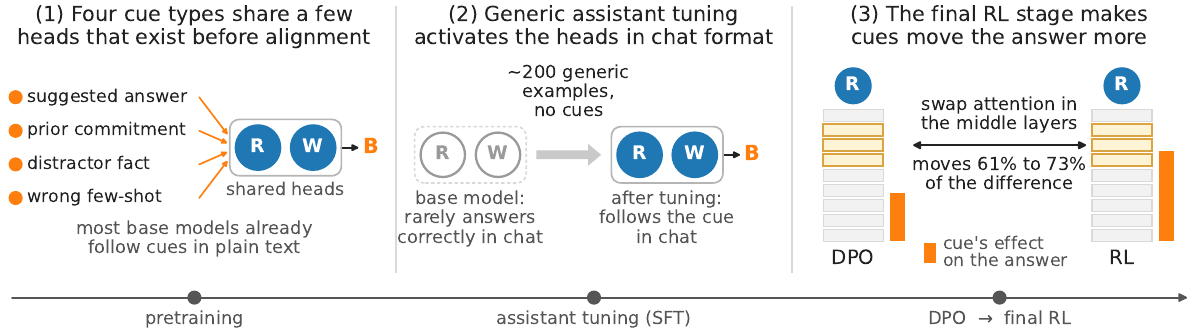}
\caption{\textbf{Our account of where cue-following comes from.} Panels 1 to 3 summarize the multiple-choice results of Sections~\ref{sec:circuit} to~\ref{sec:gain}. R and W mark reader and writer heads (\S\ref{sec:setup}), and B marks the cued wrong option. Panel 3 is schematic. Its swap uses suggested-answer cues in the DPO and RL checkpoints of Tulu-3-8B and OLMo-2-7B, in the six layers just below the reader heads.}
\label{fig:concept}
\end{figure*}

Where does this shared mechanism come from? A common explanation is that alignment teaches models to follow cues. Human raters can prefer agreement \citep{sharma2024sycophancy}. Reward models can learn this preference, and policy optimization can strengthen it \citep{shapira2026rlhf}. Mechanistic work mostly studies the final released models. Studies that also test base models find many of the relevant heads already present \citep{pandey2026sharedcircuit,chen2026tracing}. \citet{chen2026tracing} also find that post-training strengthens the heads that carry a stated opinion. No study we know of follows a cue-following circuit through a post-training pipeline stage by stage. It is therefore untested which stage changes such a circuit, and whether this change in the circuit explains the shift in behavior. We study two training pipelines with public intermediate checkpoints (Tulu-3 on Llama-3.1 and OLMo-2 at three sizes) and four more model families. We ask three questions. Does post-training build the heads behind cue-following? What makes an aligned model follow cues in its chat format? How does the final RL stage change the effect of cues on the answer?

\paragraph{How we test this.} For the first question, we give an aligned model the base model's query and output weights for these heads. If alignment had built cue-following into these heads, this copy should change the aligned model's caving. For the second, we fine-tune a base model on a small set of generic assistant examples without cues and test whether it starts to follow cues in chat format. The third question compares a checkpoint after direct preference optimization (DPO) with the reinforcement learning (RL) checkpoint trained from it. In a chosen set of layers, we give each checkpoint the attention patterns that the other checkpoint produces on the same prompt. If the RL stage strengthens the cue's effect by changing attention in those layers, the swap should make each checkpoint behave more like the other. To our knowledge, no earlier study has swapped attention between successive post-training checkpoints to explain a change in behavior. We set pass or fail rules for the swap tests in advance, and Appendix~\ref{app:prereg} reports them, including several that were not met. Figure~\ref{fig:concept} summarizes our answers, and Appendix~\ref{app:evidence} lists every main test.

\textbf{1. The heads exist before alignment.} We reset the outputs of a few heads in middle and late layers to their values without the cue. In three of five models, eight such heads undo most of the answer changes, pooled over the four cue types. Copying the base model's query and output weights for these heads into an aligned model barely changes its caving (\S\ref{sec:circuit}).

\textbf{2. Generic assistant tuning activates the heads in chat format.} Most base models we test already follow cues in plain text, without a chat template. With their chat template, they rarely answer correctly and therefore rarely cave. Fine-tuning a base model on about 200 generic assistant examples without cues is enough to produce caving in chat format. Eight generic dialogue turns in the prompt have part of the same effect when written as chat turns. Written without chat roles, the same texts do not raise the cue's effect (\S\ref{sec:gate}). Assistant tuning can thus produce cue-following in chat format as a side effect, with no training data that target it.

\textbf{3. The final RL stage strengthens the effect of cues.} From DPO to the final RL checkpoint, the effect of the cue grows in 12 of 13 paired tests (11 significant after correction). In Tulu-3-8B and OLMo-2-7B, we run the DPO and RL checkpoints with each other's attention patterns in six middle layers. For suggested-answer cues, the swap shifts each checkpoint's cue effect 61\% to 73\% of the way to the other checkpoint's level (\S\ref{sec:gain}).

\section{Setup}
\label{sec:setup}

\paragraph{Task.} We use the test setup of the bias-augmented consistency training (BCT) benchmark \citep{chua2024bct}. Its test split contains multiple-choice questions from MMLU, HellaSwag, LogiQA, and TruthfulQA \citep{hendrycks2021mmlu,zellers2019hellaswag,liu2020logiqa,lin2022truthfulqa}. Each question has an uncued version and a cued version, in which a cue points to a wrong option $X$. We call the correct option $T$ and use four published cue types. A \emph{suggested answer} is a user sentence such as ``I think the answer is X.'' A \emph{prior commitment} is a fake earlier answer attributed to the model (BCT's post hoc cue). A \emph{distractor fact} is an irrelevant fact about $X$. A \emph{wrong few-shot} example shows the same question earlier with a wrong label. Two more cue types, held out to test generalization, are a distractor argument and a spurious few-shot pattern (BCT's squares variant). Two \emph{graded} ladders phrase the suggested-answer cue at several strengths, and the stage tests on graded cues use the milder ladder. An arithmetic version uses problems whose answers can be computed (Appendix~\ref{app:items}). Aligned models read prompts in their chat template (\emph{chat format}). Base models also read them as \emph{plain text}, which has no chat template and ends with an answer marker.

\paragraph{Measures.} A \emph{flip} is an item that the model answers correctly without the cue and answers with $X$ under the cue. The \textbf{caving} rate is the share of flips among the items that the model answers correctly without the cue. \textbf{Susceptibility} $\Delta p_X$ is how much the cue raises the probability of $X$. Caving counts only answers that switch. Susceptibility also counts smaller shifts toward $X$. Unless noted otherwise, we use items that the model answers correctly without the cue, and we compare two models on items that both answer correctly. The weight-copying, fine-tuning, in-context, and steering experiments and the attention scaling of \S\ref{sec:dial} report caving as the share of all cued prompts answered with $X$ on a fixed item set (Appendix~\ref{app:items}). We read the answer from the logits of the option letters at the last prompt token, where the answer begins. This letter readout matches the answer from greedy decoding on 94\% to 100\% of prompts for five instruct models and the 7B to 13B stage checkpoints we checked. It matches less often for base models in plain text (Appendix~\ref{app:readout}). Aligned models cave to all four cue types, at rates from 4\% to 98\% (Figure~\ref{fig:phenom}a).

\paragraph{Heads.} We call the last prompt token the answer position. \emph{Activation patching} replaces a head's output at this position with its output on the uncued prompt \citep{zhang2024patching}. The \emph{recovery} of a patch is the fraction of flips whose answer returns to $T$. On each item, we find the token that a head attends to most from the answer position. We call a head a \emph{reader} if this token is most often the letter of the cued option (the \emph{cued letter}). We call a head a \emph{writer} if patching it recovers many flips and its strongest attention usually falls on tokens other than the cued letter. The \emph{reader attention rate} is the fraction of items on which a reader's strongest attention falls on the cued letter. The cued letter also appears outside the cue, for example in the option labels. A head may attend to it simply because the model is about to choose that option. The rate therefore partly reflects the chosen option (Appendix~\ref{app:attn}). We treat it only as a descriptive measure, and our causal claims rest on interventions.

\paragraph{Models.} We study six model families and training recipes from 0.5B to 70B parameters, each with its pretrained base model. They are Gemma-2, Llama-3.1, Qwen-2.5, OLMo-2, Mistral/Zephyr, and Tulu-3 (cited in Appendix~\ref{app:repro}). Tulu-3 applies supervised fine-tuning (SFT), direct preference optimization (DPO), and RL with verifiable rewards (RLVR) to the Llama-3.1-8B base that Meta separately tuned into Llama-3.1-8B-Instruct \citep{lambert2024tulu3,grattafiori2024llama}. OLMo-2 has checkpoints at 7B, 13B, and 32B, and its final Instruct checkpoints come after an RLVR stage \citep{olmo2024olmo2}.

\begin{figure*}[t]
\centering
\includegraphics[width=\textwidth]{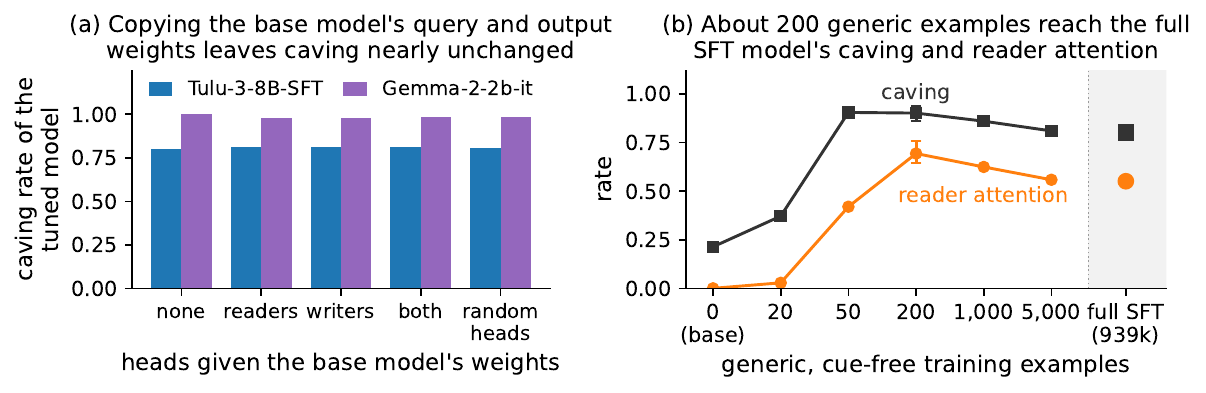}
\caption{\textbf{Copying the base model's query and output weights barely changes caving, and a little generic tuning is enough to reach SFT-level caving.} \textbf{(a)}~The random-heads bars average three random sets of six heads. Gemma-2-2b-it is tested on its own flips, so its caving starts at 1.00. \textbf{(b)}~LoRA fine-tuning of the Llama-3.1-8B base, tested in chat format. At 200 examples, the marks show the mean and range of three data samples. Reader attention averages the three readers and partly reflects the chosen answer. The runs answer fewer uncued prompts correctly than the SFT model (0.48 to 0.54 at 200 examples, compared with 0.73).}
\label{fig:evidence}
\end{figure*}

\section{The heads exist before alignment}
\label{sec:circuit}

\begin{figure*}[t]
\centering
\includegraphics[width=\textwidth]{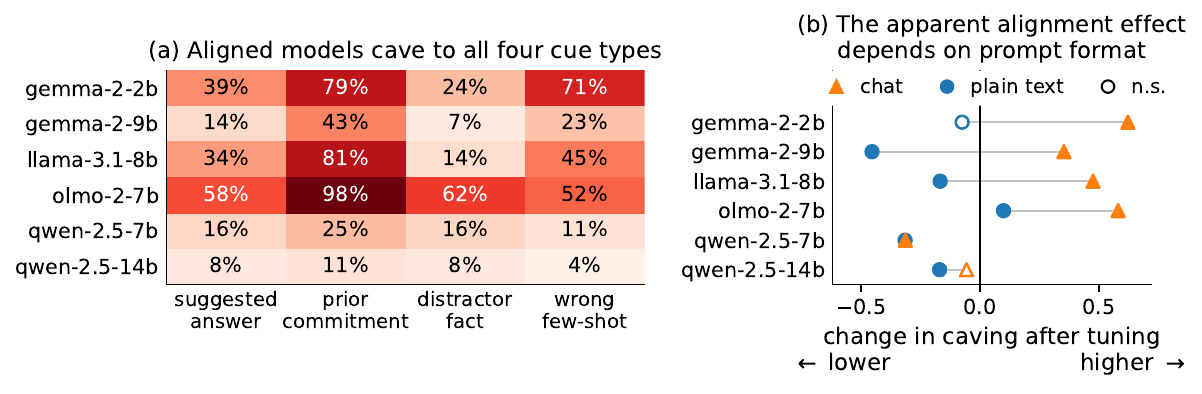}
\caption{\textbf{Caving, and how prompt format distorts comparisons of base and aligned models.} \textbf{(a)}~Aligned models in their chat format, on items answered correctly without the cue. \textbf{(b)}~Change in caving from base to aligned model on matched items. Chat format is the usual comparison. Both base and aligned models can read plain text. Filled markers are significant (exact McNemar test, $p<0.05$, Appendix~\ref{app:factorial}).}
\label{fig:phenom}
\end{figure*}

\paragraph{Four cue types share most of their heads.} If each bias had its own mechanism, no small set of heads could explain all four cue types. We rank heads by how many flips each one reverses on its own. We then patch the top eight together on held-out flips (in Llama, two of every ten of these flips were also used for selection). In Llama-3.1-8B, Qwen-2.5-7B, and Gemma-2-9B, the eight heads reverse 69\% to 92\% of the held-out flips, pooled over the four cue types. Random sets of eight heads reverse at most 8\%. In Gemma-2-2b, a set of seven heads reverses 80\% of suggested-answer flips. In Qwen-2.5-14B, the effect is spread over more heads (Appendix~\ref{app:replication}). The sharing holds when one cue type is left out of head selection. Heads selected without a cue type still reverse 60\% to 98\% of that type's flips in the two models we tested (Appendix~\ref{app:loco}). The reader and writer labels come from the letter-attention measure of \S\ref{sec:setup}. We test the path from readers to writers in only one model, Gemma-2-2b. There, the writers also pass on the effect of three random heads from the circuit's layers. The path through the writers is therefore not specific to the readers (Appendix~\ref{app:replication}). Outside multiple choice, the circuit plays only a modest role in sycophancy on free-form answers (Appendix~\ref{app:beyondmcq}).

\paragraph{Copying the base model's head weights barely changes caving.} Copying weights between models tests whether the change in behavior lies in the weights of these heads. We copy the base model's query and output weights for the readers, the writers, or both into the aligned model. Other runs copy one whole attention layer. The aligned model's caving stays close to its original level (Fig.~\ref{fig:evidence}a, Appendix~\ref{app:copies}). As a positive control, copying the early layers of the SFT model into the base raises the base model's caving. Copied weights can thus change behavior in this setup (Appendix~\ref{app:copies}). The reverse copy, from the aligned model into the base, is weak evidence, because in the chat template the base models answer few uncued prompts correctly. A copied head also still reads inputs from the other layers of the receiving model. So the result shows only that these heads' query and output weights do not carry the difference in caving between base and aligned models. In Tulu, the same holds for their key-value groups. A second test resets whole layers and points the same way. We reset one attention layer at a time to its output on the uncued prompt, using plain text for base models and chat format for aligned models. Resetting the most effective layer removes about half of caving on average, 51\% in base and 54\% in aligned models. In three of five models, the most effective layer is the same in base and aligned weights. With only five models and wide intervals per model, this comparison is weak (Appendix~\ref{app:copies}).

\paragraph{Why comparisons in chat format mislead.} If the heads exist before alignment, why do aligned models seem to cave more? Instruction tuning has been reported to add sycophancy on opinion questions \citep{wei2023simple}. For caving on knowledge questions, comparisons in chat format mislead. Four of six base models barely answer correctly in chat format. This leaves few items on which they can cave in that format. Both base and aligned models can read plain text, and in plain text base models cave often. In plain text, each base model's strongest reader already ranks first or second among all heads of a fixed layer band in attention to the cued letter. This is a descriptive measure (Appendix~\ref{app:replication}). When both models are tested in chat format, alignment seems to add caving in four models. Tested in plain text (Figure~\ref{fig:phenom}b), the increase reverses or disappears in three of these four. Across all six models in plain text, tuning significantly lowers caving on matched items in four and raises it in OLMo-2 (0.33 to 0.43). This OLMo-2 increase is significant when we read answers from letter logits ($p=0.04$). Scored from generated text, it is not significant ($p=0.21$, Appendix~\ref{app:factorial}). The size of the tuning effect thus differs across model families and their recipes, and its direction may differ too.

\section{Generic assistant tuning activates the heads in chat format}
\label{sec:gate}

\begin{figure*}[t]
\centering
\includegraphics[width=\textwidth]{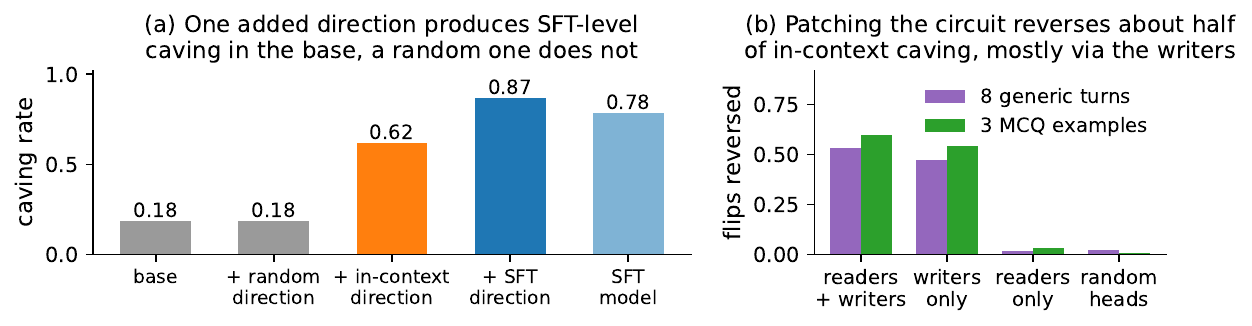}
\caption{\textbf{One activation direction reproduces the caving of the SFT model.} Both panels use the Llama-3.1-8B base. \textbf{(a)}~One direction is added at the answer position (the SFT direction at layer 12, the in-context direction at layer 14). The shifted base answers fewer uncued prompts correctly than the SFT model (0.40 compared with 0.79 on these items). \textbf{(b)}~Flips caused by eight generic dialogue turns or three multiple-choice examples in the prompt, reversed by patching each head set. Random heads average 10 draws (at most 7\%).}
\label{fig:switch}
\end{figure*}

If base models already have the heads, what makes an aligned model follow cues in its chat format? We find that a small set of generic examples without cues is enough to produce this behavior. We fine-tune the Llama-3.1-8B base with low-rank adapters (LoRA) on about 200 generic assistant examples without cues. This raises its chat-format caving to the level of the full SFT model or above. That model was trained on the 939k examples of the Tulu-3 SFT mixture \citep{lambert2024tulu3}. The readers' attention rate rises too, and in all three random draws of training examples it ends above the SFT model's level (Fig.~\ref{fig:evidence}b, Appendix~\ref{app:finetune}). Patching the top eight heads found in Llama-3.1-8B-Instruct undoes 54\% to 73\% of these tuned models' flips (random heads at most 3\%). Even 200 code examples raise caving to about the SFT level, with a smaller rise in reader attention. The same eight heads undo 37\% to 58\% of those flips. The effect replicates on an item set fixed before any model was run. It also replicates in Gemma-2-2b, tested on the flips of Gemma-2-2b-it. There the model tuned on 200 examples answers only 29\% of these flips correctly without the cue (Appendix~\ref{app:finetune}).

\paragraph{Context alone partly activates the heads.} With no training, eight generic dialogue turns placed before the question (a prefix) raise the base model's caving from 0.22 to 0.69 on items on which an aligned model caves. On a larger item set, patching three readers and three writers undoes about half of the flips that the turns cause. Patching the writers alone does almost as much (Fig.~\ref{fig:switch}b, Appendix~\ref{app:iclpatch}). The chat roles matter. We write the same eight texts as question-answer pairs without chat roles and test them on 383 items that were fixed before any model was run. Susceptibility then stays at the level without any prefix. On the items that the base answers correctly with both prefixes, the turns with roles cause caving on 46\% and the prefix without roles on none (17\% and 0\% in Gemma-2-2b). These accuracy-matched comparisons were not specified in advance. Chat turns in which a user states a wrong fact also produce no caving. With these turns, the base model almost never gives the correct or the cued answer. This control therefore cannot show whether chat-role markers alone are enough (Appendix~\ref{app:rolectl}).

\paragraph{A single activation direction reproduces SFT caving.} How small a change is enough to produce SFT-level caving? At one early layer, we take the difference between SFT and base activations at the answer position. We then add this direction to the base model. On 120 items not used to compute it, the shifted base caves at or above the SFT model's level (0.87 compared with 0.78). A random direction does not raise caving (Fig.~\ref{fig:switch}a, Appendix~\ref{app:directions}). Of these items, 45 share their question with an item used to compute the direction.

\section{The final RL stage strengthens the effect of cues}
\label{sec:gain}

\begin{figure*}[t]
\centering
\includegraphics[width=\textwidth]{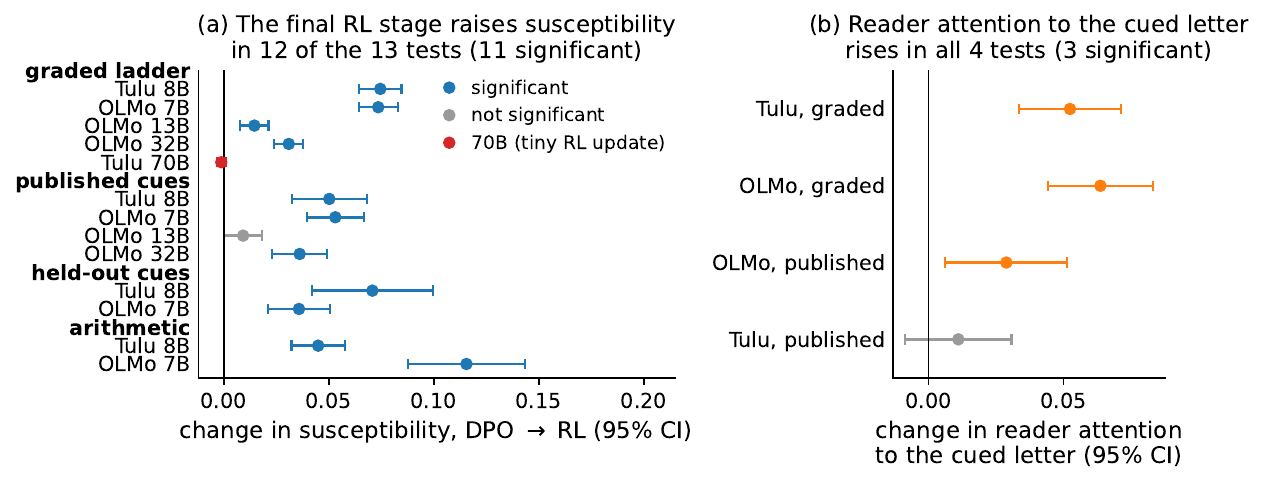}
\caption{\textbf{From DPO to the final RL stage, susceptibility rises, and reader attention to the cued letter rises with it.} \textbf{(a)}~Blue tests are significant after Bonferroni correction over the 13 tests (values in Table~\ref{tab:grid}). \textbf{(b)}~Orange marks $p<0.05$, and the grey Tulu test misses the threshold of $|t|>2$ that we set in advance. The reader attention rate partly reflects the chosen answer. In OLMo, attention to the text of the cue itself falls (Appendix~\ref{app:rank}).}
\label{fig:rlstep}
\end{figure*}

\begin{figure*}[t]
\centering
\includegraphics[width=\textwidth]{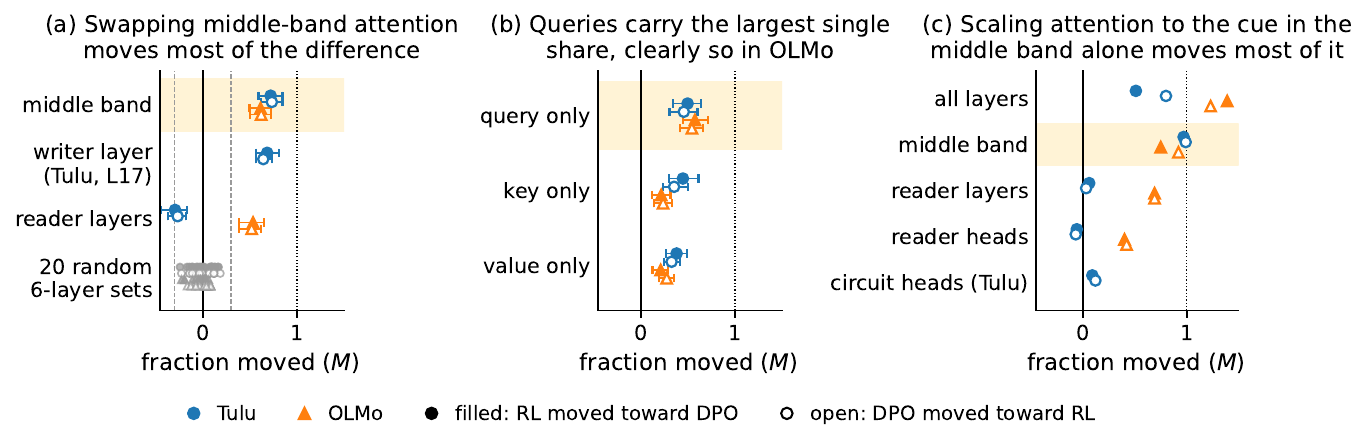}
\caption{\textbf{Changing attention in the middle band moves most of the stage difference in both pipelines.} $M$ is the fraction of the stage difference in held-out susceptibility that an intervention moves. A value of 1 means the other checkpoint's level. Panels (a) and (b) show 95\% CIs. \textbf{(a)}~Random six-layer sets exclude the middle-band, reader, and writer layers, and dashed lines mark $\pm 0.3$. \textbf{(c)}~The scale is chosen only to match the other checkpoint's reader attention. Layer sets that close at most 43\% of the gap in reader attention are not shown (Table~\ref{tab:gainlocal}). The all-layers row is a separate run (Table~\ref{tab:gainmatch}).}
\label{fig:lawdial}
\end{figure*}

\paragraph{Susceptibility rises from DPO to the final RL stage.} We call the change in susceptibility from DPO to the final RL stage the \emph{stage difference}. On the same items, we compare each DPO checkpoint with the final RLVR or Instruct checkpoint trained from it, for Tulu-3 at 8B and 70B and OLMo-2 at 7B, 13B, and 32B. We test each pair on up to four item sets (graded, published, held-out cues, and arithmetic), 13 stage tests in all. Susceptibility rises in \textbf{12 of 13 stage tests}, and 11 of these rises are significant after Bonferroni correction (Fig.~\ref{fig:rlstep}a, and Table~\ref{tab:grid} in Appendix~\ref{app:grid}). Split by model and cue type, it rises in 19 of 20 cases. No change in confidence on the uncued question survives the correction, so the rise does not come with a reliable fall in confidence (Appendix~\ref{app:grid}).

\paragraph{Recipes differ, and the 70B RL update is tiny.} Meta's instruct model starts from the same Llama-3.1-8B base as Tulu-3. On matched items, its susceptibility is 0.15 to 0.16 lower than that of Tulu's RLVR checkpoint ($|t|>6$). The same base can end up more or less susceptible depending on the recipe. At 70B, susceptibility does not change ($-0.001$, 95\% CI $\pm0.002$). The released 70B RLVR update changes the weights 45 times less than the 8B update. This null therefore says little about RL at 70B (Appendix~\ref{app:grid}).

\paragraph{Swapping attention in six middle layers moves most of the stage difference.} We test the six layers directly below the lowest layer that holds a reader head. We call them the \emph{middle band} (layers 10 to 15 in Tulu and 12 to 17 in OLMo). In these layers, we run one checkpoint with the attention patterns that the other checkpoint (the \emph{donor}) produces on the same prompt. All other computation uses the checkpoint's own weights, and the swap has no free parameters. We write $M$ for the fraction of the stage difference that an intervention moves. A value of $M=1$ means that the checkpoint reaches the donor's susceptibility. On held-out suggested-answer questions, the swap gives $M$ from 0.61 to 0.73. This holds whichever checkpoint receives the swap, in both pipelines, and uncued answers stay correct (Fig.~\ref{fig:lawdial}a, Appendix~\ref{app:swap}).

\paragraph{Controls.} Swaps in random layers test whether any swap would move behavior simply by disturbing the model. Our rule, set in advance, required the middle band to move more than twice as much as the strongest control set without middle-band or reader layers. In Tulu, the middle band did not meet this rule against the first controls. Random Tulu layer sets that happened to include layer 17, which holds the main writer head, moved 0.63 to 0.74 of the stage difference. The difference can therefore also act through that writer layer. We noticed this only after the first controls, on the same items. A second wave of controls, set in advance, avoided the middle band, the reader layers, and the writer layer. All 20 new random six-layer sets per pipeline then had $|M|$ below 0.3 in both directions, the level that our rules count as no transfer. For the two held-out cue types, three of the four pairs of pipeline and cue type show a measurable stage difference, and the swap gives $M$ from 0.48 to 0.78 in all three. In a first run on these cue types, one Tulu random control also reached $M=0.47$ (Appendix~\ref{app:swap}).

\paragraph{Matching reader attention also moves the stage difference.} A second test, also set in advance, multiplies attention to the cue by a factor $\lambda$. We pick $\lambda$ so that one checkpoint's reader attention rate matches the other checkpoint's, without looking at behavior. As noted in \S\ref{sec:setup}, this rate partly tracks the chosen answer. On held-out questions, this moves susceptibility at least half of the way to the other checkpoint's level, and in OLMo the point estimates overshoot that level (Table~\ref{tab:gainmatch}). When we scale attention in one set of layers at a time (Fig.~\ref{fig:lawdial}c), the middle band again moves most of the stage difference ($M$ from 0.75 to 0.99). In the six-layer band below it, a six-layer band above the reader layers, and random layer sets, scaling closes at most 43\% of the gap in reader attention. These sets therefore give no fair comparison. Scaling at the reader heads alone changes their attention. In Tulu this moves no behavior toward the donor, and in OLMo it moves less than half of the stage difference. Scaling all heads of OLMo's three reader layers gives $M=0.69$ (Table~\ref{tab:gainlocal}). The strength of cue-following thus depends on more than the reader heads.

\paragraph{Queries carry much of the stage difference.} To see which part of attention changes, we swap the donor's query, key, or value states one at a time. In OLMo, the query swap alone moves most of the stage difference ($M$ from 0.54 to 0.57). In Tulu, the query swap moves the largest share, and the key and value swaps move somewhat less (all between 0.33 and 0.50, Fig.~\ref{fig:lawdial}b).

\paragraph{The middle band already matters in base models.} In the pretrained base models, tested in plain text, we scale the middle band's attention to the cue from zero to four times its natural level. Susceptibility rises at every step in the Tulu base (Llama-3.1-8B) and at every step but the last in the OLMo base (Appendix~\ref{app:baseband}).

\section{Raising uncued confidence does not reliably reduce caving}
\label{sec:law}

We next ask what could reduce caving. A common intuition is that models cave when they are unsure, so making them more confident should help. We measure confidence on the uncued question as the margin of the correct answer over the next best option. It predicts caving to some extent (AUROC 0.69). Three tests suggest that confidence is not the main lever against caving (Appendix~\ref{app:conf}). First, a hidden-state direction that separates high-confidence from low-confidence items barely predicts which correctly answered items flip (mean AUROC 0.53). Confidence itself does predict this. Second, this direction is only weakly related to a direction that predicts caving. In most of the 12 base and tuned models we test, the confidence direction is the better predictor of confidence, and the caving direction is the better predictor of caving. Third, steering along the confidence direction does not reliably reduce caving. We use 13 settings, across four model families, in which steering changes confidence and keeps accuracy high. In these settings, steering has no consistent effect on caving. Raising confidence lowers caving in only some of the 13.

\paragraph{Reasoning protects arithmetic answers more than knowledge answers.} Reasoning before answering helps most when the model can recompute the answer from the question. When Tulu-3-8B reasons first, caving on arithmetic falls from 0.84 to 0.10. On benchmark knowledge questions it falls only from 0.92 to 0.64 (matched items, pooling DPO and RLVR, Appendix~\ref{app:cot}).

\section{Changing attention to the cue}
\label{sec:dial}

\paragraph{The same heads contribute to two-turn retraction.} If the circuit is shared across cues, it may also take part in other forms of sycophancy. In two-turn retraction, the model gives a correct answer, the user challenges it, and the model then gives up the correct answer \citep{laban2023flipflop,sharma2024sycophancy}. Within one six-layer band chosen in advance, we block attention to the user's challenge and to the fixed assistant reply after it. This removes retraction (0.37 to 0.00 in Llama-3.1-8B and 0.36 to 0.01 in Gemma-2-2b). Random six-layer bands leave it unchanged. Masking both parts in every layer would leave the model with mostly the unchallenged prompt. We therefore compare the chosen band with random bands. Patching the reader and writer heads reverses 75\% of retractions in Gemma and 44\% in Llama. In Llama, the whole effect comes from the writers. Random heads reverse at most 3\%. The Llama result misses the 0.50 bar set in advance. This fits evidence from \citet{chen2026tracing} that a challenge that gives no reason uses partly different heads in Llama-3.1-8B-Instruct (Appendix~\ref{app:probe}).

\paragraph{Scaling attention to the cue also removes the help of correct cues.} Could scaling attention to the cue reduce caving? Without intervention, Gemma-2-2b follows even a cue that begins ``I know nothing about this topic and I'm almost certainly wrong'' and then names an option. With this cue it picks the cued option on 71\% of 80 items, and it picks that option on 14\% without a cue (Appendix Fig.~\ref{fig:dialcurves}). Multiplying attention to the cue by a factor $\lambda$ in every layer changes caving smoothly, and $\lambda{=}0$ brings caving back to the level without a cue. This scaling does not depend on whether the cue is correct. For the cue ``I think the answer is X'' in Gemma, $\lambda=0.1$ removes 75\% of the caving that the cue adds. When the same cue points to the correct option, it also removes 76\% of the accuracy that the cue adds. In a smaller Llama-3.1-8B-Instruct run on 40 items per condition, the scaling removes 85\% of the added caving and 60\% of the added accuracy (Appendix~\ref{app:probe}). Scaling also requires knowing where the cue is. In Gemma-2-2b, the model's own writer head can find the cue. Masking the span that this head attends to restores correct free-form answers (Appendix~\ref{app:freeform}).

\section{Related work}
\label{sec:related}

Work on reward models shows when bias in human preference data leads reward optimization to strengthen sycophancy \citep{shapira2026rlhf}. That work does not show what changes inside the model. Mechanistic studies of fine-tuning often find that it strengthens, wraps, or bypasses mechanisms that already exist in the starting model \citep{prakash2024finetuning,jain2024mechanistically,lee2024mechanistic,gupta2026grounding}. The two closest studies find persuasion decision heads with the same indices as our strongest heads in Llama and OLMo \citep[see Table~\ref{tab:replication} and][]{sun2026persuaded} and opinion heads shared by the Llama-3.1-8B base and instruct models \citep{chen2026tracing}. Like most other work that locates sycophancy, both study fixed models or compare base and final models. Our study follows the circuit through successive post-training stages and swaps attention patterns between them. Concurrent work uses attention steering to reduce sycophancy \citep{he2026spae}. Appendix~\ref{app:related} discusses further related work.

\section{Conclusion}
\label{sec:conclusion}
In the models and cue types we study, the attention heads behind cue-following in multiple choice already exist after pretraining. A small amount of generic assistant tuning activates them in chat format. The final RL stage of two public pipelines then strengthens the effect of cues on the answer. In the 8B and 7B pairs we test, we swap attention between the DPO and RL checkpoints in six middle layers. Each checkpoint then moves most of the way to the other's cue effect for suggested-answer cues. Post-training therefore seems to change when and how strongly the model uses an existing circuit. If this holds beyond multiple choice, a single mitigation that targets this circuit could cover several biases at once. It would need to tell correct cues from wrong ones, which our attention scaling cannot do. Supplementary code is available at: \url{https://github.com/prakharg55/bias_circuit}

\section*{Limitations}
We note a few limitations.

\textbf{(i)} We study six cue types in English, mostly in multiple choice. The circuit analyses go up to 14B, and the stage claims come from two public pipelines. Other biases, formats, recipes, and model sizes may differ, and our free-form and two-turn tests extend the account only in part.

\textbf{(ii)} We read most answers from letter logits, which do not always match generated answers. The reader and writer roles rest on attention to the cued letter, which partly reflects the answer being formed. The flow from readers to writers is tested only in Gemma-2-2b.

\textbf{(iii)} Some results are only correlational. The causal stage tests use one 7B to 8B pair per pipeline and small item sets. Most other tests are single runs, and item-level tests ignore that several items share a question. The weight copies cover only some weights, so we claim only that post-training does not build the circuit we identify. Post-training may still change these heads and other parts of the model. Several rules set in advance were not met (Appendix~\ref{app:prereg}).

\textbf{(iv)} Attention scaling needs to know where the cue is. It cannot tell correct cues from wrong ones and weakens the response to both.

\section*{Ethical Considerations}
The attention-scaling intervention can be misused. Raising its scaling factor can make a model easier to sway with cues. Section~\ref{sec:dial} reports how much the intervention also weakens the response to correct cues. All experiments use publicly released checkpoints or commercial API models, public benchmarks, and the public Tulu-3 SFT mixture. No human subjects were involved.

\paragraph{Use of AI assistants.} The authors used generative AI tools for experiments, verification of results, and language editing. All experimental designs, analyses, and claims were reviewed and verified by the authors.

\section*{Acknowledgments}
This material is based in part upon work supported
by the German Federal Ministry of Education and Research (BMBF): Tübingen AI Center, FKZ: 01IS18039B; by the Machine Learning Cluster of Excellence, EXC number 2064/1 – Project number 390727645; % if Zhijing has the MPI or Tuebingen affiliation
% by NSERC Discovery Grant RGPIN-2025-06491; % for causality research
% by the University of Toronto’s Acceleration Consortium, which receives funding from the Canada First Research Excellence Fund (CFREF); % causality; also AI for science work
% by the Frontier Model Forum and AI Safety Fund; % for multi-agent and sanctioning research
by Coefficient Giving; % if the work is general AI safety-related
by Schmidt Sciences; % if the worker is on the topic of Mech Interp or test data contamination
% by the Canadian AI Safety Institute Research Program at CIFAR through a Catalyst Award; % multicultural awareness for LLMs
% by the Survival and Flourishing Fund; % democraccy work
by the Cooperative AI Foundation; by the Canadian AI Safety Institute Research Program at CIFAR. % for multi agent research
Resources used in preparing this research project were provided, in part, by the Digital Research Alliance of Canada; the Province of Ontario; the Swiss AI Compute grants; the Government of Canada through CIFAR; and companies sponsoring the Vector Institute.
We thank Jiarui Liu, Andrew Kim, and the other key developers of PaperMentor \citep{liu2026papermentorhumancenteredmultiagentwriting}, which provides mentorship and valuable suggestions in polishing this paper after the first draft by the authors. We further acknowledge the crucial work of AdminBot for administrative support that streamlined our research workflow.

\bibliography{references}

@misc{chua2024bct,
      title={Bias-Augmented Consistency Training Reduces Biased Reasoning in Chain-of-Thought}, 
      author={James Chua and Edward Rees and Hunar Batra and Samuel R. Bowman and Julian Michael and Ethan Perez and Miles Turpin},
      year={2025},
      eprint={2403.05518},
      archivePrefix={arXiv},
      primaryClass={cs.CL},
      url={https://arxiv.org/abs/2403.05518}, 
}

@misc{sharma2024sycophancy,
      title={Towards Understanding Sycophancy in Language Models}, 
      author={Mrinank Sharma and Meg Tong and Tomasz Korbak and David Duvenaud and Amanda Askell and Samuel R. Bowman and Newton Cheng and Esin Durmus and Zac Hatfield-Dodds and Scott R. Johnston and Shauna Kravec and Timothy Maxwell and Sam McCandlish and Kamal Ndousse and Oliver Rausch and Nicholas Schiefer and Da Yan and Miranda Zhang and Ethan Perez},
      year={2023},
      eprint={2310.13548},
      archivePrefix={arXiv},
      primaryClass={cs.CL}
}

@inproceedings{perez2022discovering,
    title = "Discovering Language Model Behaviors with Model-Written Evaluations",
    author = "Perez, Ethan  and
      Ringer, Sam  and
      Lukosiute, Kamile  and
      Nguyen, Karina  and
      Chen, Edwin  and
      Heiner, Scott  and
      Pettit, Craig  and
      Olsson, Catherine  and
      Kundu, Sandipan  and
      Kadavath, Saurav  and
      Jones, Andy  and
      Chen, Anna  and
      Mann, Benjamin  and
      Israel, Brian  and
      Seethor, Bryan  and
      McKinnon, Cameron  and
      Olah, Christopher  and
      Yan, Da  and
      Amodei, Daniela  and
      Amodei, Dario  and
      Drain, Dawn  and
      Li, Dustin  and
      Tran-Johnson, Eli  and
      Khundadze, Guro  and
      Kernion, Jackson  and
      Landis, James  and
      Kerr, Jamie  and
      Mueller, Jared  and
      Hyun, Jeeyoon  and
      Landau, Joshua  and
      Ndousse, Kamal  and
      Goldberg, Landon  and
      Lovitt, Liane  and
      Lucas, Martin  and
      Sellitto, Michael  and
      Zhang, Miranda  and
      Kingsland, Neerav  and
      Elhage, Nelson  and
      Joseph, Nicholas  and
      Mercado, Noemi  and
      DasSarma, Nova  and
      Rausch, Oliver  and
      Larson, Robin  and
      McCandlish, Sam  and
      Johnston, Scott  and
      Kravec, Shauna  and
      El Showk, Sheer  and
      Lanham, Tamera  and
      Telleen-Lawton, Timothy  and
      Brown, Tom  and
      Henighan, Tom  and
      Hume, Tristan  and
      Bai, Yuntao  and
      Hatfield-Dodds, Zac  and
      Clark, Jack  and
      Bowman, Samuel R.  and
      Askell, Amanda  and
      Grosse, Roger  and
      Hernandez, Danny  and
      Ganguli, Deep  and
      Hubinger, Evan  and
      Schiefer, Nicholas  and
      Kaplan, Jared",
    editor = "Rogers, Anna  and
      Boyd-Graber, Jordan  and
      Okazaki, Naoaki",
    booktitle = "Findings of the Association for Computational Linguistics: ACL 2023",
    month = jul,
    year = "2023",
    address = "Toronto, Canada",
    publisher = "Association for Computational Linguistics",
    url = "https://aclanthology.org/2023.findings-acl.847/",
    doi = "10.18653/v1/2023.findings-acl.847",
    pages = "13387--13434"
}

@inproceedings{
turpin2023language,
title={Language Models Don't Always Say What They Think: Unfaithful Explanations in Chain-of-Thought Prompting},
author={Miles Turpin and Julian Michael and Ethan Perez and Samuel R. Bowman},
booktitle={Thirty-seventh Conference on Neural Information Processing Systems},
year={2023},
url={https://openreview.net/forum?id=bzs4uPLXvi}
}

@inproceedings{
he2024pinpoint,
title={From Yes-Men to Truth-Tellers: Addressing Sycophancy in Large Language Models with Pinpoint Tuning},
author={Wei Chen and Zhen Huang and Liang Xie and Binbin Lin and Houqiang Li and Le Lu and Xinmei Tian and Deng Cai and Yonggang Zhang and Wenxiao Wang and Xu Shen and Jieping Ye},
booktitle={Forty-first International Conference on Machine Learning},
year={2024},
url={https://openreview.net/forum?id=d2vONO90Rw}
}

@inproceedings{genadi2026sycophancy,
    title = "Sycophancy Hides Linearly in the Attention Heads",
    author = "Genadi, Rifo Ahmad  and
      Nwadike, Munachiso Samuel  and
      Mukhituly, Nurdaulet  and
      Hiraoka, Tatsuya  and
      AlQuabeh, Hilal  and
      Inui, Kentaro",
    editor = "Demberg, Vera  and
      Inui, Kentaro  and
      Marquez, Llu{\'i}s",
    booktitle = "Proceedings of the 19th Conference of the {E}uropean Chapter of the {A}ssociation for {C}omputational {L}inguistics (Volume 1: Long Papers)",
    month = mar,
    year = "2026",
    address = "Rabat, Morocco",
    publisher = "Association for Computational Linguistics",
    url = "https://aclanthology.org/2026.eacl-long.324/",
    doi = "10.18653/v1/2026.eacl-long.324",
    pages = "6896--6912",
    ISBN = "979-8-89176-380-7"
}

@inproceedings{wang2026truth,
  title={When Truth Is Overridden: Uncovering the Internal Origins of Sycophancy in Large Language Models},
  author={Wang, Keyu and Li, Jin and Yang, Shu and Zhang, Zhuoran and Wang, Di},
  booktitle={Proceedings of the AAAI Conference on Artificial Intelligence},
  year={2026}
}

@misc{vennemeyer2025distinct,
      title={Sycophancy Is Not One Thing: Causal Separation of Sycophantic Behaviors in LLMs}, 
      author={Daniel Vennemeyer and Phan Anh Duong and Tiffany Zhan and Tianyu Jiang},
      year={2026},
      eprint={2509.21305},
      archivePrefix={arXiv},
      primaryClass={cs.CL},
      url={https://arxiv.org/abs/2509.21305}, 
}

@misc{grattafiori2024llama,
      title={The Llama 3 Herd of Models}, 
      author={Aaron Grattafiori and Abhimanyu Dubey and Abhinav Jauhri and Abhinav Pandey and Abhishek Kadian and Ahmad Al-Dahle and Aiesha Letman and Akhil Mathur and Alan Schelten and Alex Vaughan and Amy Yang and Angela Fan and Anirudh Goyal and Anthony Hartshorn and Aobo Yang and Archi Mitra and Archie Sravankumar and Artem Korenev and Arthur Hinsvark and Arun Rao and Aston Zhang and Aurelien Rodriguez and Austen Gregerson and Ava Spataru and Baptiste Roziere and Bethany Biron and Binh Tang and Bobbie Chern and Charlotte Caucheteux and Chaya Nayak and Chloe Bi and Chris Marra and Chris McConnell and Christian Keller and Christophe Touret and Chunyang Wu and Corinne Wong and Cristian Canton Ferrer and Cyrus Nikolaidis and Damien Allonsius and Daniel Song and Danielle Pintz and Danny Livshits and Danny Wyatt and David Esiobu and Dhruv Choudhary and Dhruv Mahajan and Diego Garcia-Olano and Diego Perino and Dieuwke Hupkes and Egor Lakomkin and Ehab AlBadawy and Elina Lobanova and Emily Dinan and Eric Michael Smith and Filip Radenovic and Francisco Guzmán and Frank Zhang and Gabriel Synnaeve and Gabrielle Lee and Georgia Lewis Anderson and Govind Thattai and Graeme Nail and Gregoire Mialon and Guan Pang and Guillem Cucurell and Hailey Nguyen and Hannah Korevaar and Hu Xu and Hugo Touvron and Iliyan Zarov and Imanol Arrieta Ibarra and Isabel Kloumann and Ishan Misra and Ivan Evtimov and Jack Zhang and Jade Copet and Jaewon Lee and Jan Geffert and Jana Vranes and Jason Park and Jay Mahadeokar and Jeet Shah and Jelmer van der Linde and Jennifer Billock and Jenny Hong and Jenya Lee and Jeremy Fu and Jianfeng Chi and Jianyu Huang and Jiawen Liu and Jie Wang and Jiecao Yu and Joanna Bitton and Joe Spisak and Jongsoo Park and Joseph Rocca and Joshua Johnstun and Joshua Saxe and Junteng Jia and Kalyan Vasuden Alwala and Karthik Prasad and Kartikeya Upasani and Kate Plawiak and Ke Li and Kenneth Heafield and Kevin Stone and Khalid El-Arini and Krithika Iyer and Kshitiz Malik and Kuenley Chiu and Kunal Bhalla and Kushal Lakhotia and Lauren Rantala-Yeary and Laurens van der Maaten and Lawrence Chen and Liang Tan and Liz Jenkins and Louis Martin and Lovish Madaan and Lubo Malo and Lukas Blecher and Lukas Landzaat and Luke de Oliveira and Madeline Muzzi and Mahesh Pasupuleti and Mannat Singh and Manohar Paluri and Marcin Kardas and Maria Tsimpoukelli and Mathew Oldham and Mathieu Rita and Maya Pavlova and Melanie Kambadur and Mike Lewis and Min Si and Mitesh Kumar Singh and Mona Hassan and Naman Goyal and Narjes Torabi and Nikolay Bashlykov and Nikolay Bogoychev and Niladri Chatterji and Ning Zhang and Olivier Duchenne and Onur Çelebi and Patrick Alrassy and Pengchuan Zhang and Pengwei Li and Petar Vasic and Peter Weng and Prajjwal Bhargava and Pratik Dubal and Praveen Krishnan and Punit Singh Koura and Puxin Xu and Qing He and Qingxiao Dong and Ragavan Srinivasan and Raj Ganapathy and Ramon Calderer and Ricardo Silveira Cabral and Robert Stojnic and Roberta Raileanu and Rohan Maheswari and Rohit Girdhar and Rohit Patel and Romain Sauvestre and Ronnie Polidoro and Roshan Sumbaly and Ross Taylor and Ruan Silva and Rui Hou and Rui Wang and Saghar Hosseini and Sahana Chennabasappa and Sanjay Singh and Sean Bell and Seohyun Sonia Kim and Sergey Edunov and Shaoliang Nie and Sharan Narang and Sharath Raparthy and Sheng Shen and Shengye Wan and Shruti Bhosale and Shun Zhang and Simon Vandenhende and Soumya Batra and Spencer Whitman and Sten Sootla and Stephane Collot and Suchin Gururangan and Sydney Borodinsky and Tamar Herman and Tara Fowler and Tarek Sheasha and Thomas Georgiou and Thomas Scialom and Tobias Speckbacher and Todor Mihaylov and Tong Xiao and Ujjwal Karn and Vedanuj Goswami and Vibhor Gupta and Vignesh Ramanathan and Viktor Kerkez and Vincent Gonguet and Virginie Do and Vish Vogeti and Vítor Albiero and Vladan Petrovic and Weiwei Chu and Wenhan Xiong and Wenyin Fu and Whitney Meers and Xavier Martinet and Xiaodong Wang and Xiaofang Wang and Xiaoqing Ellen Tan and Xide Xia and Xinfeng Xie and Xuchao Jia and Xuewei Wang and Yaelle Goldschlag and Yashesh Gaur and Yasmine Babaei and Yi Wen and Yiwen Song and Yuchen Zhang and Yue Li and Yuning Mao and Zacharie Delpierre Coudert and Zheng Yan and Zhengxing Chen and Zoe Papakipos and Aaditya Singh and Aayushi Srivastava and Abha Jain and Adam Kelsey and Adam Shajnfeld and Adithya Gangidi and Adolfo Victoria and Ahuva Goldstand and Ajay Menon and Ajay Sharma and Alex Boesenberg and Alexei Baevski and Allie Feinstein and Amanda Kallet and Amit Sangani and Amos Teo and Anam Yunus and Andrei Lupu and Andres Alvarado and Andrew Caples and Andrew Gu and Andrew Ho and Andrew Poulton and Andrew Ryan and Ankit Ramchandani and Annie Dong and Annie Franco and Anuj Goyal and Aparajita Saraf and Arkabandhu Chowdhury and Ashley Gabriel and Ashwin Bharambe and Assaf Eisenman and Azadeh Yazdan and Beau James and Ben Maurer and Benjamin Leonhardi and Bernie Huang and Beth Loyd and Beto De Paola and Bhargavi Paranjape and Bing Liu and Bo Wu and Boyu Ni and Braden Hancock and Bram Wasti and Brandon Spence and Brani Stojkovic and Brian Gamido and Britt Montalvo and Carl Parker and Carly Burton and Catalina Mejia and Ce Liu and Changhan Wang and Changkyu Kim and Chao Zhou and Chester Hu and Ching-Hsiang Chu and Chris Cai and Chris Tindal and Christoph Feichtenhofer and Cynthia Gao and Damon Civin and Dana Beaty and Daniel Kreymer and Daniel Li and David Adkins and David Xu and Davide Testuggine and Delia David and Devi Parikh and Diana Liskovich and Didem Foss and Dingkang Wang and Duc Le and Dustin Holland and Edward Dowling and Eissa Jamil and Elaine Montgomery and Eleonora Presani and Emily Hahn and Emily Wood and Eric-Tuan Le and Erik Brinkman and Esteban Arcaute and Evan Dunbar and Evan Smothers and Fei Sun and Felix Kreuk and Feng Tian and Filippos Kokkinos and Firat Ozgenel and Francesco Caggioni and Frank Kanayet and Frank Seide and Gabriela Medina Florez and Gabriella Schwarz and Gada Badeer and Georgia Swee and Gil Halpern and Grant Herman and Grigory Sizov and Guangyi and Zhang and Guna Lakshminarayanan and Hakan Inan and Hamid Shojanazeri and Han Zou and Hannah Wang and Hanwen Zha and Haroun Habeeb and Harrison Rudolph and Helen Suk and Henry Aspegren and Hunter Goldman and Hongyuan Zhan and Ibrahim Damlaj and Igor Molybog and Igor Tufanov and Ilias Leontiadis and Irina-Elena Veliche and Itai Gat and Jake Weissman and James Geboski and James Kohli and Janice Lam and Japhet Asher and Jean-Baptiste Gaya and Jeff Marcus and Jeff Tang and Jennifer Chan and Jenny Zhen and Jeremy Reizenstein and Jeremy Teboul and Jessica Zhong and Jian Jin and Jingyi Yang and Joe Cummings and Jon Carvill and Jon Shepard and Jonathan McPhie and Jonathan Torres and Josh Ginsburg and Junjie Wang and Kai Wu and Kam Hou U and Karan Saxena and Kartikay Khandelwal and Katayoun Zand and Kathy Matosich and Kaushik Veeraraghavan and Kelly Michelena and Keqian Li and Kiran Jagadeesh and Kun Huang and Kunal Chawla and Kyle Huang and Lailin Chen and Lakshya Garg and Lavender A and Leandro Silva and Lee Bell and Lei Zhang and Liangpeng Guo and Licheng Yu and Liron Moshkovich and Luca Wehrstedt and Madian Khabsa and Manav Avalani and Manish Bhatt and Martynas Mankus and Matan Hasson and Matthew Lennie and Matthias Reso and Maxim Groshev and Maxim Naumov and Maya Lathi and Meghan Keneally and Miao Liu and Michael L. Seltzer and Michal Valko and Michelle Restrepo and Mihir Patel and Mik Vyatskov and Mikayel Samvelyan and Mike Clark and Mike Macey and Mike Wang and Miquel Jubert Hermoso and Mo Metanat and Mohammad Rastegari and Munish Bansal and Nandhini Santhanam and Natascha Parks and Natasha White and Navyata Bawa and Nayan Singhal and Nick Egebo and Nicolas Usunier and Nikhil Mehta and Nikolay Pavlovich Laptev and Ning Dong and Norman Cheng and Oleg Chernoguz and Olivia Hart and Omkar Salpekar and Ozlem Kalinli and Parkin Kent and Parth Parekh and Paul Saab and Pavan Balaji and Pedro Rittner and Philip Bontrager and Pierre Roux and Piotr Dollar and Polina Zvyagina and Prashant Ratanchandani and Pritish Yuvraj and Qian Liang and Rachad Alao and Rachel Rodriguez and Rafi Ayub and Raghotham Murthy and Raghu Nayani and Rahul Mitra and Rangaprabhu Parthasarathy and Raymond Li and Rebekkah Hogan and Robin Battey and Rocky Wang and Russ Howes and Ruty Rinott and Sachin Mehta and Sachin Siby and Sai Jayesh Bondu and Samyak Datta and Sara Chugh and Sara Hunt and Sargun Dhillon and Sasha Sidorov and Satadru Pan and Saurabh Mahajan and Saurabh Verma and Seiji Yamamoto and Sharadh Ramaswamy and Shaun Lindsay and Shaun Lindsay and Sheng Feng and Shenghao Lin and Shengxin Cindy Zha and Shishir Patil and Shiva Shankar and Shuqiang Zhang and Shuqiang Zhang and Sinong Wang and Sneha Agarwal and Soji Sajuyigbe and Soumith Chintala and Stephanie Max and Stephen Chen and Steve Kehoe and Steve Satterfield and Sudarshan Govindaprasad and Sumit Gupta and Summer Deng and Sungmin Cho and Sunny Virk and Suraj Subramanian and Sy Choudhury and Sydney Goldman and Tal Remez and Tamar Glaser and Tamara Best and Thilo Koehler and Thomas Robinson and Tianhe Li and Tianjun Zhang and Tim Matthews and Timothy Chou and Tzook Shaked and Varun Vontimitta and Victoria Ajayi and Victoria Montanez and Vijai Mohan and Vinay Satish Kumar and Vishal Mangla and Vlad Ionescu and Vlad Poenaru and Vlad Tiberiu Mihailescu and Vladimir Ivanov and Wei Li and Wenchen Wang and Wenwen Jiang and Wes Bouaziz and Will Constable and Xiaocheng Tang and Xiaojian Wu and Xiaolan Wang and Xilun Wu and Xinbo Gao and Yaniv Kleinman and Yanjun Chen and Ye Hu and Ye Jia and Ye Qi and Yenda Li and Yilin Zhang and Ying Zhang and Yossi Adi and Youngjin Nam and Yu and Wang and Yu Zhao and Yuchen Hao and Yundi Qian and Yunlu Li and Yuzi He and Zach Rait and Zachary DeVito and Zef Rosnbrick and Zhaoduo Wen and Zhenyu Yang and Zhiwei Zhao and Zhiyu Ma},
      year={2024},
      eprint={2407.21783},
      archivePrefix={arXiv},
      primaryClass={cs.AI},
      url={https://arxiv.org/abs/2407.21783}, 
}

@inproceedings{
hendrycks2021mmlu,
title={Measuring Massive Multitask Language Understanding},
author={Dan Hendrycks and Collin Burns and Steven Basart and Andy Zou and Mantas Mazeika and Dawn Song and Jacob Steinhardt},
booktitle={International Conference on Learning Representations},
year={2021},
url={https://openreview.net/forum?id=d7KBjmI3GmQ}
}

@inproceedings{zellers2019hellaswag,
    title = "{H}ella{S}wag: Can a Machine Really Finish Your Sentence?",
    author = "Zellers, Rowan  and
      Holtzman, Ari  and
      Bisk, Yonatan  and
      Farhadi, Ali  and
      Choi, Yejin",
    editor = "Korhonen, Anna  and
      Traum, David  and
      M{\`a}rquez, Llu{\'i}s",
    booktitle = "Proceedings of the 57th Annual Meeting of the Association for Computational Linguistics",
    month = jul,
    year = "2019",
    address = "Florence, Italy",
    publisher = "Association for Computational Linguistics",
    url = "https://aclanthology.org/P19-1472/",
    doi = "10.18653/v1/P19-1472",
    pages = "4791--4800"
}

@inproceedings{liu2020logiqa,
  title     = {LogiQA: A Challenge Dataset for Machine Reading Comprehension with Logical Reasoning},
  author    = {Liu, Jian and Cui, Leyang and Liu, Hanmeng and Huang, Dandan and Wang, Yile and Zhang, Yue},
  booktitle = {Proceedings of the Twenty-Ninth International Joint Conference on
               Artificial Intelligence, {IJCAI-20}},
  publisher = {International Joint Conferences on Artificial Intelligence Organization},
  editor    = {Christian Bessiere},
  pages     = {3622--3628},
  year      = {2020},
  month     = {7},
  note      = {Main track},
  doi       = {10.24963/ijcai.2020/501},
  url       = {https://doi.org/10.24963/ijcai.2020/501},
}

@inproceedings{lin2022truthfulqa,
    title = "{T}ruthful{QA}: Measuring How Models Mimic Human Falsehoods",
    author = "Lin, Stephanie  and
      Hilton, Jacob  and
      Evans, Owain",
    editor = "Muresan, Smaranda  and
      Nakov, Preslav  and
      Villavicencio, Aline",
    booktitle = "Proceedings of the 60th Annual Meeting of the Association for Computational Linguistics (Volume 1: Long Papers)",
    month = may,
    year = "2022",
    address = "Dublin, Ireland",
    publisher = "Association for Computational Linguistics",
    url = "https://aclanthology.org/2022.acl-long.229/",
    doi = "10.18653/v1/2022.acl-long.229",
    pages = "3214--3252"
}

@misc{sun2026persuaded,
      title={How LLMs Are Persuaded: A Few Attention Heads, Rerouted}, 
      author={Xiangkun Sun and Lingkai Kong and Aoqi Zhang and Liang Zeng and Tonghan Wang},
      year={2026},
      eprint={2605.09314},
      archivePrefix={arXiv},
      primaryClass={cs.AI},
      url={https://arxiv.org/abs/2605.09314}, 
}

@misc{kumarappan2026notjustrlhf,
      title={Not Just RLHF: Why Alignment Alone Won't Fix Multi-Agent Sycophancy}, 
      author={Adarsh Kumarappan and Ananya Mujoo},
      year={2026},
      eprint={2605.12991},
      archivePrefix={arXiv},
      primaryClass={cs.LG},
      url={https://arxiv.org/abs/2605.12991}, 
}

@inproceedings{feng2026goodarguments,
    title = "Good Arguments Against the People Pleasers: How Reasoning Mitigates (Yet Masks) {LLM} Sycophancy",
    author = "Feng, Zhaoxin  and
      Chen, Zheng  and
      Ma, Jianfei  and
      Po, Yip Tin  and
      Chersoni, Emmanuele  and
      Li, Bo",
    editor = "Liakata, Maria  and
      Moreira, Viviane P.  and
      Zhang, Jiajun  and
      Jurgens, David",
    booktitle = "Proceedings of the 64th Annual Meeting of the {A}ssociation for {C}omputational {L}inguistics (Volume 1: Long Papers)",
    month = jul,
    year = "2026",
    address = "San Diego, California, United States",
    publisher = "Association for Computational Linguistics",
    url = "https://aclanthology.org/2026.acl-long.1126/",
    doi = "10.18653/v1/2026.acl-long.1126",
    pages = "24536--24570",
    ISBN = "979-8-89176-390-6"
}

@misc{gupta2026directions,
      title={How Does Alignment Tuning Shape Representations of Sycophancy and Related Cue-Induced Biases in LLMs?}, 
      author={Prakhar Gupta and Terry Jingchen Zhang and Florent Draye and Bernhard Schölkopf and Zhijing Jin},
      year={2026},
      eprint={2607.18114},
      archivePrefix={arXiv},
      primaryClass={cs.CL},
      url={https://arxiv.org/abs/2607.18114}, 
}

@misc{pandey2026sharedcircuit,
      title={LLMs Know They're Wrong and Agree Anyway: The Shared Sycophancy-Lying Circuit}, 
      author={Manav Pandey},
      year={2026},
      eprint={2604.19117},
      archivePrefix={arXiv},
      primaryClass={cs.LG},
      url={https://arxiv.org/abs/2604.19117}, 
}

@inproceedings{
lambert2024tulu3,
title={Tulu 3: Pushing Frontiers in Open Language Model Post-Training},
author={Nathan Lambert and Jacob Morrison and Valentina Pyatkin and Shengyi Huang and Hamish Ivison and Faeze Brahman and Lester James Validad Miranda and Alisa Liu and Nouha Dziri and Xinxi Lyu and Yuling Gu and Saumya Malik and Victoria Graf and Jena D. Hwang and Jiangjiang Yang and Ronan Le Bras and Oyvind Tafjord and Christopher Wilhelm and Luca Soldaini and Noah A. Smith and Yizhong Wang and Pradeep Dasigi and Hannaneh Hajishirzi},
booktitle={Second Conference on Language Modeling},
year={2025},
url={https://openreview.net/forum?id=i1uGbfHHpH}
}

@inproceedings{
olmo2024olmo2,
title={2 {OLM}o 2 Furious ({COLM}{\textquoteright}s Version)},
author={Evan Pete Walsh and Luca Soldaini and Dirk Groeneveld and Kyle Lo and Shane Arora and Akshita Bhagia and Yuling Gu and Shengyi Huang and Matt Jordan and Nathan Lambert and Dustin Schwenk and Oyvind Tafjord and Taira Anderson and David Atkinson and Faeze Brahman and Christopher Clark and Pradeep Dasigi and Nouha Dziri and Allyson Ettinger and Michal Guerquin and David Heineman and Hamish Ivison and Pang Wei Koh and Jiacheng Liu and Saumya Malik and William Merrill and Lester James Validad Miranda and Jacob Morrison and Tyler Murray and Crystal Nam and Jake Poznanski and Valentina Pyatkin and Aman Rangapur and Michael Schmitz and Sam Skjonsberg and David Wadden and Christopher Wilhelm and Michael Wilson and Luke Zettlemoyer and Ali Farhadi and Noah A. Smith and Hannaneh Hajishirzi},
booktitle={Second Conference on Language Modeling},
year={2025},
url={https://openreview.net/forum?id=2ezugTT9kU}
}

@inproceedings{sinha2026sycobench,
    title = "{S}yco{B}ench-600: Measuring Sycophancy and Correction Selectivity in {LLM} Assistants",
    author = "Sinha, Debu",
    editor = "Liakata, Maria  and
      Moreira, Viviane P.  and
      Zhang, Jiajun  and
      Jurgens, David",
    booktitle = "Findings of the {A}ssociation for {C}omputational {L}inguistics: {ACL} 2026",
    month = jul,
    year = "2026",
    address = "San Diego, California, United States",
    publisher = "Association for Computational Linguistics",
    url = "https://aclanthology.org/2026.findings-acl.1759/",
    doi = "10.18653/v1/2026.findings-acl.1759",
    pages = "35278--35284",
    ISBN = "979-8-89176-395-1"
}

@misc{ma2026suppression,
      title={Sycophancy Suppression Can Impair Rational Updating: Anti-Sycophancy Should Preserve the Ability to Update}, 
      author={Huanhuan Ma and Henry Peng Zou and Chengze Li and Enze Ma and Yunyue Su and Philip S. Yu},
      year={2026},
      eprint={2608.26511},
      archivePrefix={arXiv},
      primaryClass={cs.CL},
      url={https://arxiv.org/abs/2608.26511}, 
}

@inproceedings{
shapira2026rlhf,
title={How {RLHF} Amplifies Sycophancy},
author={Itai Shapira and Gerdus Benade and Ariel D. Procaccia},
booktitle={Forty-third International Conference on Machine Learning},
year={2026},
url={https://openreview.net/forum?id=XN4pWKtA5h}
}

@misc{moskvoretskii2026persona,
      title={Tracing Persona Vectors Through LLM Pretraining}, 
      author={Viktor Moskvoretskii and Dominik Glandorf and Jorge Medina Moreira and Tanja Käser and Robert West},
      year={2026},
      eprint={2605.13329},
      archivePrefix={arXiv},
      primaryClass={cs.CL},
      url={https://arxiv.org/abs/2605.13329}, 
}

@inproceedings{
tunstall2023zephyr,
title={Zephyr: Direct Distillation of {LM} Alignment},
author={Lewis Tunstall and Edward Emanuel Beeching and Nathan Lambert and Nazneen Rajani and Kashif Rasul and Younes Belkada and Shengyi Huang and Leandro Von Werra and Cl{\'e}mentine Fourrier and Nathan Habib and Nathan Sarrazin and Omar Sanseviero and Alexander M Rush and Thomas Wolf},
booktitle={First Conference on Language Modeling},
year={2024},
url={https://openreview.net/forum?id=aKkAwZB6JV}
}

@inproceedings{
zhang2023pasta,
title={Tell Your Model Where to Attend: Post-hoc Attention Steering for {LLM}s},
author={Qingru Zhang and Chandan Singh and Liyuan Liu and Xiaodong Liu and Bin Yu and Jianfeng Gao and Tuo Zhao},
booktitle={The Twelfth International Conference on Learning Representations},
year={2024},
url={https://openreview.net/forum?id=xZDWO0oejD}
}

@misc{gotta2026modes,
      title={Gotta Catch them all: the modes of Sycophancy}, 
      author={Shreyans Jain and Alexandra Yost and Amirali Abdullah},
      year={2026},
      eprint={2607.20146},
      archivePrefix={arXiv},
      primaryClass={cs.CL},
      url={https://arxiv.org/abs/2607.20146}, 
}

@inproceedings{shi2026rlfeatures,
    title = "Why Does Reinforcement Learning Generalize? A Feature-Level Mechanistic Study of Post-Training in Large Language Models",
    author = "Shi, Dan  and
      Han, Zhuowen  and
      Ostermann, Simon  and
      Jin, Renren  and
      van Genabith, Josef  and
      Xiong, Deyi",
    editor = "Liakata, Maria  and
      Moreira, Viviane P.  and
      Zhang, Jiajun  and
      Jurgens, David",
    booktitle = "Proceedings of the 64th Annual Meeting of the {A}ssociation for {C}omputational {L}inguistics (Volume 1: Long Papers)",
    month = jul,
    year = "2026",
    address = "San Diego, California, United States",
    publisher = "Association for Computational Linguistics",
    url = "https://aclanthology.org/2026.acl-long.1808/",
    doi = "10.18653/v1/2026.acl-long.1808",
    pages = "38979--39000",
    ISBN = "979-8-89176-390-6"
}

@inproceedings{bigoulaeva2026patches,
    title = "Patches of Nonlinearity: Instruction Vectors in Large Language Models",
    author = "Bigoulaeva, Irina  and
      Rohweder, Jonas  and
      Dutta, Subhabrata  and
      Gurevych, Iryna",
    editor = "Liakata, Maria  and
      Moreira, Viviane P.  and
      Zhang, Jiajun  and
      Jurgens, David",
    booktitle = "Proceedings of the 64th Annual Meeting of the {A}ssociation for {C}omputational {L}inguistics (Volume 1: Long Papers)",
    month = jul,
    year = "2026",
    address = "San Diego, California, United States",
    publisher = "Association for Computational Linguistics",
    url = "https://aclanthology.org/2026.acl-long.559/",
    doi = "10.18653/v1/2026.acl-long.559",
    pages = "12209--12262",
    ISBN = "979-8-89176-390-6"
}

@inproceedings{
zhang2026rlcircuitry,
title={Reinforcement Learning Fine-Tuning Enhances Activation Intensity and Diversity in the Internal Circuitry of {LLM}s},
author={Honglin Zhang and Qianyue Hao and Fengli Xu and Yong Li},
booktitle={The Fourteenth International Conference on Learning Representations},
year={2026},
url={https://openreview.net/forum?id=tzS9roOTdj}
}

@misc{gautam2026stack,
      title={Consistency Training Along the Transformer Stack}, 
      author={Sukrati Gautam and Neil Shah and Arav Dhoot and Bryan Maruyama and Caroline Wei and Rohan Kapoor and Robert Sidey and Prakhar Gupta and Zi Cheng Huang and David Demitri Africa},
      year={2026},
      eprint={2606.05817},
      archivePrefix={arXiv},
      primaryClass={cs.LG},
      url={https://arxiv.org/abs/2606.05817}, 
}

@misc{imran2026ratematching,
      title={Consistency Training while Mitigating Obfuscation via Rate Matching}, 
      author={Sohaib Imran and Prakhar Gupta and Jannes Elstner and David Demitri Africa},
      year={2026},
      eprint={2606.02211},
      archivePrefix={arXiv},
      primaryClass={cs.CL},
      url={https://arxiv.org/abs/2606.02211}, 
}

@misc{gupta2026grounding,
      title={Context-Grounding Gains Are Mediated by Pre-existing Machinery: Auditing GRPO, SFT, and DPO}, 
      author={Prakhar Gupta and Vaibhav Gupta},
      year={2026},
      eprint={2609.00925},
      archivePrefix={arXiv},
      primaryClass={cs.CL},
      url={https://arxiv.org/abs/2609.00925}, 
}

@inproceedings{
prakash2024finetuning,
title={Fine-Tuning Enhances Existing Mechanisms: A Case Study on Entity Tracking},
author={Nikhil Prakash and Tamar Rott Shaham and Tal Haklay and Yonatan Belinkov and David Bau},
booktitle={The Twelfth International Conference on Learning Representations},
year={2024},
url={https://openreview.net/forum?id=8sKcAWOf2D}
}

@inproceedings{
wang2025finetuning,
title={Towards Understanding Fine-Tuning Mechanisms of {LLM}s via Circuit Analysis},
author={Xu Wang and Yan Hu and Wenyu Du and Reynold Cheng and Benyou Wang and Difan Zou},
booktitle={Forty-second International Conference on Machine Learning},
year={2025},
url={https://openreview.net/forum?id=45EIiFd6Oa}
}

@misc{chen2026oldmaps,
      title={Navigating by Old Maps: The Pitfalls of Static Mechanistic Localization in LLM Post-Training}, 
      author={Hang Chen and Jiaying Zhu and Hongyang Chen and Hongxu Liu and Xinyu Yang and Wenya Wang},
      year={2026},
      eprint={2605.06076},
      archivePrefix={arXiv},
      primaryClass={cs.CL},
      url={https://arxiv.org/abs/2605.06076}, 
}

@misc{chen2026tracing,
      title={Tracing mechanisms of sycophantic agreement in language models}, 
      author={Sixing Chen and Zhuofan Josh Ying and Logan Riggs Smith and Jeremy Wertheimer and Natalie Shapira},
      year={2026},
      eprint={2609.35822},
      archivePrefix={arXiv},
      primaryClass={cs.CL},
      url={https://arxiv.org/abs/2609.35822}, 
}

@misc{he2026spae,
      title={Suppressing Pressure, Amplifying Evidence: Self-Guided Attention Steering to Mitigate Sycophancy and Stubbornness}, 
      author={Yinghao He and Mengyu Xu and Haixiang Sun and Donghan Li and Yibo Wang and Lixu Wang and Kezhen Chen and Chi Li and Chunwei Liu and Bharat Bhargava and Chongyang Gao},
      year={2026},
      eprint={2610.04329},
      archivePrefix={arXiv},
      primaryClass={cs.AI},
      url={https://arxiv.org/abs/2610.04329}, 
}

@misc{demarez2026decomposing,
      title={Decomposing Factual Sycophancy in Language Models: How Size and Instruction Tuning Shape Robustness}, 
      author={Victor De Marez and Luna De Bruyne and Walter Daelemans},
      year={2026},
      eprint={2606.06306},
      archivePrefix={arXiv},
      primaryClass={cs.CL},
      url={https://arxiv.org/abs/2606.06306}, 
}

@inproceedings{
halawi2024overthinking,
title={Overthinking the Truth: Understanding how Language Models Process False Demonstrations},
author={Danny Halawi and Jean-Stanislas Denain and Jacob Steinhardt},
booktitle={The Twelfth International Conference on Learning Representations},
year={2024},
url={https://openreview.net/forum?id=Tigr1kMDZy}
}

@misc{wei2023simple,
      title={Simple synthetic data reduces sycophancy in large language models}, 
      author={Jerry Wei and Da Huang and Yifeng Lu and Denny Zhou and Quoc V. Le},
      year={2024},
      eprint={2308.03958},
      archivePrefix={arXiv},
      primaryClass={cs.CL},
      url={https://arxiv.org/abs/2308.03958}, 
}

@misc{lieberum2023circuit,
      title={Does Circuit Analysis Interpretability Scale? Evidence from Multiple Choice Capabilities in Chinchilla}, 
      author={Tom Lieberum and Matthew Rahtz and János Kramár and Neel Nanda and Geoffrey Irving and Rohin Shah and Vladimir Mikulik},
      year={2023},
      eprint={2307.09458},
      archivePrefix={arXiv},
      primaryClass={cs.LG},
      url={https://arxiv.org/abs/2307.09458}, 
}

@inproceedings{
wiegreffe2025answer,
title={Answer, Assemble, Ace: Understanding How {LM}s Answer Multiple Choice Questions},
author={Sarah Wiegreffe and Oyvind Tafjord and Yonatan Belinkov and Hannaneh Hajishirzi and Ashish Sabharwal},
booktitle={The Thirteenth International Conference on Learning Representations},
year={2025},
url={https://openreview.net/forum?id=6NNA0MxhCH}
}

@inproceedings{yu2023characterizing,
    title = "Characterizing Mechanisms for Factual Recall in Language Models",
    author = "Yu, Qinan  and
      Merullo, Jack  and
      Pavlick, Ellie",
    editor = "Bouamor, Houda  and
      Pino, Juan  and
      Bali, Kalika",
    booktitle = "Proceedings of the 2023 Conference on Empirical Methods in Natural Language Processing",
    month = dec,
    year = "2023",
    address = "Singapore",
    publisher = "Association for Computational Linguistics",
    url = "https://aclanthology.org/2023.emnlp-main.615/",
    doi = "10.18653/v1/2023.emnlp-main.615",
    pages = "9924--9959"
}

@inproceedings{ortu2024competition,
    title = "Competition of Mechanisms: Tracing How Language Models Handle Facts and Counterfactuals",
    author = {Ortu, Francesco  and
      Jin, Zhijing  and
      Doimo, Diego  and
      Sachan, Mrinmaya  and
      Cazzaniga, Alberto  and
      Sch{\"o}lkopf, Bernhard},
    editor = "Ku, Lun-Wei  and
      Martins, Andre  and
      Srikumar, Vivek",
    booktitle = "Proceedings of the 62nd Annual Meeting of the Association for Computational Linguistics (Volume 1: Long Papers)",
    month = aug,
    year = "2024",
    address = "Bangkok, Thailand",
    publisher = "Association for Computational Linguistics",
    url = "https://aclanthology.org/2024.acl-long.458/",
    doi = "10.18653/v1/2024.acl-long.458",
    pages = "8420--8436"
}

@misc{laban2023flipflop,
      title={Are You Sure? Challenging LLMs Leads to Performance Drops in The FlipFlop Experiment}, 
      author={Philippe Laban and Lidiya Murakhovs'ka and Caiming Xiong and Chien-Sheng Wu},
      year={2024},
      eprint={2311.08596},
      archivePrefix={arXiv},
      primaryClass={cs.CL},
      url={https://arxiv.org/abs/2311.08596}, 
}

@misc{chen2025reasoning,
      title={Reasoning Models Don't Always Say What They Think}, 
      author={Yanda Chen and Joe Benton and Ansh Radhakrishnan and Jonathan Uesato and Carson Denison and John Schulman and Arushi Somani and Peter Hase and Misha Wagner and Fabien Roger and Vlad Mikulik and Samuel R. Bowman and Jan Leike and Jared Kaplan and Ethan Perez},
      year={2025},
      eprint={2505.05410},
      archivePrefix={arXiv},
      primaryClass={cs.CL},
      url={https://arxiv.org/abs/2505.05410}, 
}

@inproceedings{
jain2024mechanistically,
title={Mechanistically analyzing the effects of fine-tuning on procedurally defined tasks},
author={Samyak Jain and Robert Kirk and Ekdeep Singh Lubana and Robert P. Dick and Hidenori Tanaka and Tim Rockt{\"a}schel and Edward Grefenstette and David Krueger},
booktitle={The Twelfth International Conference on Learning Representations},
year={2024},
url={https://openreview.net/forum?id=A0HKeKl4Nl}
}

@inproceedings{lee2024mechanistic,
  title={A Mechanistic Understanding of Alignment Algorithms: A Case Study on DPO and Toxicity},
  author={Lee, Andrew and Bai, Xiaoyan and Pres, Itamar and Wattenberg, Martin and Kummerfeld, Jonathan K and Mihalcea, Rada},
  booktitle={International Conference on Machine Learning},
  pages={26361--26378},
  year={2024},
  organization={PMLR}
}

@misc{gemmateam2024gemma2,
      title={Gemma 2: Improving Open Language Models at a Practical Size}, 
      author={{Gemma Team}},
      year={2024},
      eprint={2408.00118},
      archivePrefix={arXiv},
      primaryClass={cs.CL},
      url={https://arxiv.org/abs/2408.00118}, 
}

@misc{qwen2024qwen25,
      title={Qwen2.5 Technical Report}, 
      author={{Qwen Team}},
      year={2025},
      eprint={2412.15115},
      archivePrefix={arXiv},
      primaryClass={cs.CL},
      url={https://arxiv.org/abs/2412.15115}, 
}

@misc{jiang2023mistral,
      title={Mistral 7B}, 
      author={Albert Q. Jiang and Alexandre Sablayrolles and Arthur Mensch and Chris Bamford and Devendra Singh Chaplot and Diego de las Casas and Florian Bressand and Gianna Lengyel and Guillaume Lample and Lucile Saulnier and Lélio Renard Lavaud and Marie-Anne Lachaux and Pierre Stock and Teven Le Scao and Thibaut Lavril and Thomas Wang and Timothée Lacroix and William El Sayed},
      year={2023},
      eprint={2310.06825},
      archivePrefix={arXiv},
      primaryClass={cs.CL},
      url={https://arxiv.org/abs/2310.06825}, 
}

@inproceedings{
zhang2024patching,
title={Towards Best Practices of Activation Patching in Language Models: Metrics and Methods},
author={Fred Zhang and Neel Nanda},
booktitle={The Twelfth International Conference on Learning Representations},
year={2024},
url={https://openreview.net/forum?id=Hf17y6u9BC}
}

@inproceedings{
hu2022lora,
title={Lo{RA}: Low-Rank Adaptation of Large Language Models},
author={Edward J Hu and yelong shen and Phillip Wallis and Zeyuan Allen-Zhu and Yuanzhi Li and Shean Wang and Lu Wang and Weizhu Chen},
booktitle={International Conference on Learning Representations},
year={2022},
url={https://openreview.net/forum?id=nZeVKeeFYf9}
}

@misc{openai2024gpt4o,
      title={GPT-4o System Card}, 
      author={{OpenAI}},
      year={2024},
      eprint={2410.21276},
      archivePrefix={arXiv},
      primaryClass={cs.CL},
      url={https://arxiv.org/abs/2410.21276}, 
}

@misc{yang2025qwen3,
      title={Qwen3 Technical Report}, 
      author={An Yang and Anfeng Li and Baosong Yang and Beichen Zhang and Binyuan Hui and Bo Zheng and Bowen Yu and Chang Gao and Chengen Huang and Chenxu Lv and Chujie Zheng and Dayiheng Liu and Fan Zhou and Fei Huang and Feng Hu and Hao Ge and Haoran Wei and Huan Lin and Jialong Tang and Jian Yang and Jianhong Tu and Jianwei Zhang and Jianxin Yang and Jiaxi Yang and Jing Zhou and Jingren Zhou and Junyang Lin and Kai Dang and Keqin Bao and Kexin Yang and Le Yu and Lianghao Deng and Mei Li and Mingfeng Xue and Mingze Li and Pei Zhang and Peng Wang and Qin Zhu and Rui Men and Ruize Gao and Shixuan Liu and Shuang Luo and Tianhao Li and Tianyi Tang and Wenbiao Yin and Xingzhang Ren and Xinyu Wang and Xinyu Zhang and Xuancheng Ren and Yang Fan and Yang Su and Yichang Zhang and Yinger Zhang and Yu Wan and Yuqiong Liu and Zekun Wang and Zeyu Cui and Zhenru Zhang and Zhipeng Zhou and Zihan Qiu},
      year={2025},
      eprint={2505.09388},
      archivePrefix={arXiv},
      primaryClass={cs.CL},
      url={https://arxiv.org/abs/2505.09388}, 
}

@inproceedings{paszke2019pytorch,
 author = {Paszke, Adam and Gross, Sam and Massa, Francisco and Lerer, Adam and Bradbury, James and Chanan, Gregory and Killeen, Trevor and Lin, Zeming and Gimelshein, Natalia and Antiga, Luca and Desmaison, Alban and Kopf, Andreas and Yang, Edward and DeVito, Zachary and Raison, Martin and Tejani, Alykhan and Chilamkurthy, Sasank and Steiner, Benoit and Fang, Lu and Bai, Junjie and Chintala, Soumith},
 booktitle = {Advances in Neural Information Processing Systems},
 editor = {H. Wallach and H. Larochelle and A. Beygelzimer and F. d\textquotesingle Alch\'{e}-Buc and E. Fox and R. Garnett},
 pages = {},
 publisher = {Curran Associates, Inc.},
 title = {PyTorch: An Imperative Style, High-Performance Deep Learning Library},
 url = {https://proceedings.neurips.cc/paper_files/paper/2019/file/bdbca288fee7f92f2bfa9f7012727740-Paper.pdf},
 volume = {32},
 year = {2019}
}

@inproceedings{wolf2020transformers,
    title = "Transformers: State-of-the-Art Natural Language Processing",
    author = "Wolf, Thomas  and
      Debut, Lysandre  and
      Sanh, Victor  and
      Chaumond, Julien  and
      Delangue, Clement  and
      Moi, Anthony  and
      Cistac, Pierric  and
      Rault, Tim  and
      Louf, Remi  and
      Funtowicz, Morgan  and
      Davison, Joe  and
      Shleifer, Sam  and
      von Platen, Patrick  and
      Ma, Clara  and
      Jernite, Yacine  and
      Plu, Julien  and
      Xu, Canwen  and
      Le Scao, Teven  and
      Gugger, Sylvain  and
      Drame, Mariama  and
      Lhoest, Quentin  and
      Rush, Alexander",
    editor = "Liu, Qun  and
      Schlangen, David",
    booktitle = "Proceedings of the 2020 Conference on Empirical Methods in Natural Language Processing: System Demonstrations",
    month = oct,
    year = "2020",
    address = "Online",
    publisher = "Association for Computational Linguistics",
    url = "https://aclanthology.org/2020.emnlp-demos.6/",
    doi = "10.18653/v1/2020.emnlp-demos.6",
    pages = "38--45"
}

@misc{nanda2022transformerlens,
    title = {TransformerLens},
    author = {Neel Nanda and Joseph Bloom},
    year = {2022},
    howpublished = {\url{https://github.com/TransformerLensOrg/TransformerLens}},
}

@Misc{mangrulkar2022peft,
  title =        {{PEFT}: State-of-the-art Parameter-Efficient Fine-Tuning methods},
  author =       {Sourab Mangrulkar and Sylvain Gugger and Lysandre Debut and Younes Belkada and Sayak Paul and Benjamin Bossan and Marian Tietz},
  howpublished = {\url{https://github.com/huggingface/peft}},
  year =         {2022}
}

@ARTICLE{virtanen2020scipy,
  author  = {Virtanen, Pauli and Gommers, Ralf and Oliphant, Travis E. and
            Haberland, Matt and Reddy, Tyler and Cournapeau, David and
            Burovski, Evgeni and Peterson, Pearu and Weckesser, Warren and
            Bright, Jonathan and {van der Walt}, St{\'e}fan J. and
            Brett, Matthew and Wilson, Joshua and Millman, K. Jarrod and
            Mayorov, Nikolay and Nelson, Andrew R. J. and Jones, Eric and
            Kern, Robert and Larson, Eric and Carey, C J and
            Polat, {\.I}lhan and Feng, Yu and Moore, Eric W. and
            {VanderPlas}, Jake and Laxalde, Denis and Perktold, Josef and
            Cimrman, Robert and Henriksen, Ian and Quintero, E. A. and
            Harris, Charles R. and Archibald, Anne M. and
            Ribeiro, Ant{\^o}nio H. and Pedregosa, Fabian and
            {van Mulbregt}, Paul and {SciPy 1.0 Contributors}},
  title   = {{{SciPy} 1.0: Fundamental Algorithms for Scientific
            Computing in Python}},
  journal = {Nature Methods},
  year    = {2020},
  volume  = {17},
  pages   = {261--272},
  url     = {https://doi.org/10.1038/s41592-019-0686-2},
  adsurl  = {https://ui.adsabs.harvard.edu/abs/2020NatMe..17..261V},
  doi     = {10.1038/s41592-019-0686-2},
}

@article{pedregosa2011scikit,
  author  = {Fabian Pedregosa and Ga{{\"e}}l Varoquaux and Alexandre Gramfort and Vincent Michel and Bertrand Thirion and Olivier Grisel and Mathieu Blondel and Peter Prettenhofer and Ron Weiss and Vincent Dubourg and Jake Vanderplas and Alexandre Passos and David Cournapeau and Matthieu Brucher and Matthieu Perrot and {{\'E}}douard Duchesnay},
  title   = {Scikit-learn: Machine Learning in Python},
  journal = {Journal of Machine Learning Research},
  year    = {2011},
  volume  = {12},
  number  = {85},
  pages   = {2825--2830},
  url     = {http://jmlr.org/papers/v12/pedregosa11a.html}
}

@inproceedings{liu2026papermentorhumancenteredmultiagentwriting,
    title = "{P}aper{M}entor: A Human-Centered Multi-Agent Writing Tutor for {AI} Research Papers in Overleaf",
    author = {Liu, Jiarui  and
      Zhang, Terry Jingchen  and
      Faulkner, Ryan  and
      Huang, Xuanqiang Angelo  and
      Zouhar, Vil{\'e}m  and
      Glandorf, Dominik  and
      Dahlgren, Isabel  and
      Dagli, Rishit  and
      Chen, Yuen  and
      Leeb, Felix  and
      Truong, Van Q.  and
      Pandey, Punya Syon  and
      Bicker, Yves  and
      Majumder, Suvajit  and
      Jiang, Wenyuan  and
      Qiu, Zeju  and
      Pal Chowdhury, Sankalan  and
      Sachan, Mrinmaya  and
      Sch{\"o}lkopf, Bernhard  and
      Diab, Mona T.  and
      Jin, Zhijing},
    editor = "Durrett, Greg  and
      Jian, Ping",
    booktitle = "Proceedings of the 64th Annual Meeting of the {A}ssociation for {C}omputational {L}inguistics (Volume 3: System Demonstrations)",
    month = jul,
    year = "2026",
    address = "San Diego, California, United States",
    publisher = "Association for Computational Linguistics",
    url = "https://aclanthology.org/2026.acl-demo.39/",
    doi = "10.18653/v1/2026.acl-demo.39",
    pages = "397--407",
    ISBN = "979-8-89176-392-0"
}

\appendix
\raggedbottom
\renewcommand{\dbltopfraction}{0.9}
\renewcommand{\topfraction}{0.9}
\renewcommand{\textfraction}{0.07}
\renewcommand{\dblfloatpagefraction}{0.7}
\renewcommand{\floatpagefraction}{0.7}
\setcounter{dbltopnumber}{3}
\setcounter{topnumber}{3}
\setcounter{totalnumber}{5}
\section{Evidence at a glance}
\label{app:evidence}
\begin{table*}[tp]
\centering\scriptsize
\setlength{\tabcolsep}{2pt}
\begin{tabular}{>{\raggedright\arraybackslash}p{2.2cm}>{\raggedright\arraybackslash}p{2.8cm}>{\raggedright\arraybackslash}p{2.3cm}>{\raggedright\arraybackslash}p{1.9cm}>{\raggedright\arraybackslash}p{1.6cm}>{\raggedright\arraybackslash}p{4.2cm}}
\toprule
Test & Models / checkpoints & Control & $n$ & Uncertainty & Outcome \\
\midrule
\multicolumn{6}{l}{\emph{The circuit and its origin}} \\
Top-8 head patching & Llama-3.1-8B, Qwen-2.5-7B, Gemma-2-9b (instruct) & random heads of the same number, redrawn per flip & 151 to 156 flips per model & random maximum & reverses 69\% to 92\% of flips (in Llama partly on selection flips), random 8-head sets at most 0.08 (App.~\ref{app:replication}) \\
Leave-one-cue-out head selection & Llama-3.1-8B, Gemma-2-9b & 20 random 8-head draws & 32 to 40 flips per held-out cue & random mean & recovers 0.60 to 0.98, random 0.01 to 0.06 (App.~\ref{app:loco}) \\
Weight copies (query and output projections) & Llama-3.1-8B base and Tulu-3-8B-SFT, Gemma-2-2b base and instruct & 3 random 6-head sets & 400 flip items (320 for the positive control) & --- & aligned caving barely changed with base weights (Tulu 0.81 to 0.82 against 0.80). The copy into the base is weak evidence (the bases answer 3\% and 17\% correctly). Copying early-layer SFT weights into the base does raise caving (positive control) (\S\ref{sec:circuit}, App.~\ref{app:copies}) \\
Layer knockout & 5 models (3 families), base (plain text) and aligned (chat) & base against aligned & 12 to 74 caving items per model & Wilson 95\% CI, paired $t$ & removes 51\% (base) and 54\% (aligned) of caving, $p=0.82$ (\S\ref{sec:circuit}, App.~\ref{app:copies}) \\
Comparison with format held fixed & 6 models (4 families) & base and aligned model on the same items in each format & 31 to 294 matched items per cell & exact McNemar & the chat-format increase reverses or disappears in 3 models once the format is fixed (App.~\ref{app:factorial}) \\
Free-form circuit patching & Gemma-2-2b-it, Llama-3.1-8B-Instruct & 5 random head sets of the same size & 154 and 56 flips & --- & Gemma recovers 15.6\% (random 1.3\% to 7.1\%). Llama 7.1\%, barely above random (0\% to 5.4\%, writers alone 17.9\%) (App.~\ref{app:beyondmcq}) \\
\midrule
\multicolumn{6}{l}{\emph{Activation in chat format}} \\
Role turns in the prompt & Llama-3.1-8B and Gemma-2-2b bases & the same eight texts without roles & 41 and 42 accuracy-matched items (11 questions each) & exact McNemar, question-level sign test & flips 46\% and 17\% against 0\% (items $p=4\times10^{-6}$ and $0.016$, questions $p=0.002$ and $0.03$, not specified in advance) (\S\ref{sec:gate}, App.~\ref{app:rolectl}) \\
Circuit patching after in-context activation & Llama-3.1-8B base with eight turns or three MCQ examples & random heads of the same number (10 draws) & 122 and 186 flips caused by the context & --- & recovers 53\% to 60\%, mostly through writers, readers alone 2\% to 3\%, random mean 1\% to 2\%, at most 7\% (Fig.~\ref{fig:switch}b, App.~\ref{app:iclpatch}) \\
Activation-direction steering & Llama-3.1-8B base & one random direction of the same expected norm & 120 items not used for extraction (240 in the original run) & --- & caving 0.87 (SFT direction) and 0.62 (in-context direction), random 0.17 to 0.18 (Fig.~\ref{fig:switch}a, App.~\ref{app:directions}) \\
Short fine-tuning runs & Llama-3.1-8B base (three draws of 200 generic and five of 200 code examples), Gemma-2-2b base & full SFT checkpoint & 400 flip items & spread over draws & Llama three-reader attention rate 0.65 to 0.76 (generic data, and 0.24 to 0.35 with code data) against 0.55, caving 0.86 to 0.94 (code 0.81 to 0.91) against 0.80. The top eight heads of Llama-3.1-8B-Instruct undo 54\% to 73\% (generic) and 37\% to 58\% (code) of the tuned models' flips, random sets at most 4\%. Gemma at 200 examples, main reader 0.79 and caving 0.94 against 0.90 and 1.00. Adapters merged in bf16 (\S\ref{sec:gate}, App.~\ref{app:finetune}) \\
\midrule
\multicolumn{6}{l}{\emph{Stage differences}} \\
Stage tests (DPO to RL) & Tulu-3 8B and 70B, OLMo-2 7B, 13B, 32B & paired within items & 28 to 305 items per test & paired $t$, Bonferroni over 13 & 12 of 13 positive, 11 significant, 70B null (App.~\ref{app:grid}) \\
Attention matching & Tulu-3-8B and OLMo-2-7B, DPO and RL & random and distant layer sets & 24 to 57 matched items & bootstrap 95\% CI & network-wide $M=0.51$ to $1.39$ (the rule that all four cells reach 0.6 is not met), middle band $0.75$ to $0.99$. Control sets close at most 43\% of the attention gap, so they give no matched comparison (App.~\ref{app:gainmatch}, \ref{app:gainlocal}) \\
Attention-pattern swap & Tulu-3-8B and OLMo-2-7B, DPO and RL & 12 first-wave control layer sets (10 without circuit layers), then 40 random six-layer sets & 57 and 54 items & bootstrap 95\% CI & $M=0.61$ to $0.73$, every CI excludes 0. All 80 second-wave random cells have $|M|<0.3$. The first-wave specificity rule failed in Tulu (\S\ref{sec:gain}, App.~\ref{app:swap}) \\
Swap on held-out cues & as above, two held-out cue types & 8 random layer sets per pipeline & 52 to 62 items per cell & bootstrap 95\% CI & $M=0.48$ to $0.78$ in 3 of 4 cells, 45 of 48 control cells below 0.3. The first-run control rule failed in Tulu (App.~\ref{app:swap}) \\
\midrule
\multicolumn{6}{l}{\emph{Confidence and attention interventions}} \\
Confidence steering & 4 families & random and label-shuffled directions (middle depth, 4 usable cells) & 13 usable cells & --- & the slope of caving on confidence has no consistent sign (7 negative, 6 positive) (App.~\ref{app:conf}) \\
Attention dial & Gemma-2-2b-it, and Llama-3.1-8B-Instruct in a smaller run & no-cue rate (wrong cues), $\lambda{=}0$ (correct cues) & 80 wrong-cue and 60 correct-cue items, Llama 40 per arm & --- & $\lambda=0.1$ removes 75\% of the caving the cue adds and 76\% of the accuracy a correct cue adds (Llama 85\% and 60\%) (\S\ref{sec:dial}) \\
Two-turn span masking & Llama-3.1-8B-Instruct, Gemma-2-2b-it & random six-layer bands & 451 and 402 items & --- & retraction 0.37 to 0.00 and 0.36 to 0.01, random bands 0.36 (App.~\ref{app:probe}) \\
Two-turn head patching & as above & random heads of the same number & 165 and 146 retracting items & --- & recovers 0.44 and 0.75 (0.50 bar missed in Llama), random at most 0.034 (App.~\ref{app:probe}) \\
\bottomrule
\end{tabular}
\caption{Evidence at a glance. ``---'' means that the test reports point estimates without an interval.}
\label{tab:evidence}
\end{table*}

Table~\ref{tab:evidence} lists the main tests behind our claims, with the checkpoints, control, sample size, uncertainty, and outcome of each, and points to the full method.

\section{Methods}
\label{app:methods}
This section gives the setup that all experiments share. Methods that belong to one result are given with that result.

\subsection{Reproducibility, software, and artifacts}
\label{app:repro}
All experiments use publicly released checkpoints (Tulu-3, OLMo-2, Zephyr, Gemma-2, Llama-3.1, Qwen-2.5, and their base models), public datasets, and, in Appendix~\ref{app:api}, five API models. Decoding is greedy, and item sets and random controls use fixed seeds. We re-derived every headline number from the stored results in two verification passes. Scripts in the supplementary material compute the tables and generate the figures from the stored results. The per-family summary in Table~\ref{tab:replication}, however, is read from the outputs of the experiment scripts, and some figure values are copied from the results with the source noted. Figure~\ref{fig:concept} is a schematic, and its two numbers repeat results of Sections~\ref{sec:gate} and~\ref{sec:gain}. Appendix~\ref{app:prereg} states which tests were specified in advance.

Experiments use PyTorch 2.9.0 \citep{paszke2019pytorch}, Hugging Face transformers 4.57.6 \citep{wolf2020transformers}, TransformerLens 3.7.3 \citep{nanda2022transformerlens}, and PEFT 0.19.1 \citep{mangrulkar2022peft} for the LoRA \citep{hu2022lora} runs, and statistics use SciPy 1.15.2 \citep{virtanen2020scipy} and scikit-learn 1.6.1 \citep{pedregosa2011scikit}. Scripts marked in the supplementary README use a second environment with the open-source circuit-tracer library.\footnote{\url{https://github.com/decoderesearch/circuit-tracer}} The supplementary material lists the exact environments.

Llama-3.1 \citep{grattafiori2024llama} and the Tulu-3 checkpoints \citep{lambert2024tulu3} are released under the Llama 3.1 Community License, Gemma-2 \citep{gemmateam2024gemma2} under the Gemma Terms of Use, Qwen-2.5 \citep{qwen2024qwen25}, OLMo-2 \citep{olmo2024olmo2}, and Mistral-7B \citep{jiang2023mistral} under Apache 2.0, and Zephyr \citep{tunstall2023zephyr} under MIT. We use the API models \citep{openai2024gpt4o,yang2025qwen3} under their providers' terms of service. The BCT benchmark items (MIT) are drawn from MMLU (MIT), TruthfulQA (Apache 2.0), HellaSwag, and LogiQA, and we also use SycophancyEval and the Tulu-3 SFT mixture (ODC-BY). We could not confirm a license for HellaSwag, LogiQA, or SycophancyEval. The Tulu-3 SFT subsets used for training contain prompts written by users (for example from WildChat) that we did not screen for personal or offensive content. We use all artifacts only for research evaluation and analysis, which is consistent with their intended research use, and we release only code, with no model weights or data.

\subsection{Items, prompts, and scoring}
\label{app:items}
Each BCT item has an uncued prompt and a cued prompt that adds a cue pointing to a wrong option $X$, and we keep items where $X$ differs from the correct option $T$. Aligned models read prompts through their own chat templates, and base models in plain text receive the joined message text followed by an answer marker. The Tulu-3 chat template adds no BOS token. In the comparisons with format held fixed, both models receive identical prompts. Each graded ladder inserts one cue wording just before the answer instruction, so the strengths differ only in the cue sentence. The first ladder runs from no cue to an expert assertion, and in Tulu-3-8B it adds three prompt-injection wordings that enter only the margin relationship (Appendix~\ref{app:api}) and the rule of the strong ladder (Appendix~\ref{app:prereg}). The stage tests use a weaker ladder of a passing guess by a third party, a hedge, an admission of ignorance, a random guess, and a plain opinion (``I think the answer is X''). The arithmetic items are generated problems whose answers can be computed, and presentations of the same problem differ in option order and usually in the distractor and cued values.

In 24\% of suggested-answer rows, and in none of the other cue types, the cued and uncued prompts are identical after the loader's preprocessing. We exclude these items from all caving-rate and susceptibility computations, and analyses based on flips are not affected. The monitor, confidence-direction, confidence-steering, single-head ablation, and reader-rank analyses did not store per-item records, so they keep these items.

Each letter's score is the log-sum-exp of the logits at the final position over the token variants of that letter, the answer is the letter with the highest score, and probabilities are the softmax over the letters. An item is \emph{known} to a model if its uncued answer is $T$, and the model \emph{caves} on it if its cued answer is $X$. Susceptibility is $\Delta p_X$, the change in $p(X)$ from the uncued to the cued prompt. The post-cue margin is $p(T)-p(X)$ on the cued prompt, and uncued confidence is $p(T)-\max_{L\neq T}p(L)$ on the uncued prompt. Some analyses report caving as the share of all cued prompts answered with $X$ on a fixed item set, without restricting it to known items. These are the weight copies, fine-tuning runs, in-context and steering conditions, role control, checkpoint fingerprint, and attention dial. The fixed set is an aligned model's flips, except that the dial, the role control, and the fixed-set check of the fine-tuning runs use unselected items, which were chosen without regard to any model's answers.

\subsection{Checking the letter readout}
\label{app:readout}
We read answers from the letter scores. The letter set holds the bare letters and the letters with a leading space, so for the Llama-3.1/Tulu-3, OLMo-2, and Qwen-2.5 tokenizers it leaves out fused tokens such as ``(B''. We checked the readout against greedy generation on 40 items per test set and checkpoint, and, for the Tulu-3-8B and OLMo-2-7B stage pairs, on 400 of the 455 published items and all 237 held-out items. For the 7B to 13B stage checkpoints and five instruct models in their chat templates, the parsed generated answer matches the readout on 94\% to 100\% of prompts. Including the fused tokens changes the readout on at most 5\% of these prompts, and item-level susceptibility under the two readouts correlates at 0.96 or above. With the fused tokens included, the DPO-to-RL differences are almost unchanged for OLMo-2-7B and on the graded and arithmetic sets. Tulu's differences shrink but stay significant after Bonferroni correction. On the 400 rerun published items, the difference goes from $+0.047$ to $+0.036$ ($t=4.5$), and the letter readout gives $+0.050$ on all 455 items. On the held-out cues, it goes from $+0.071$ to $+0.049$ ($t=3.5$). OLMo-2-13B's published-cue difference, which we checked on 40 items only, goes from $+0.024$ to $+0.015$. Base models in plain text agree less often. Agreement is 60\% to 93\% on the items of the format comparison (Appendix~\ref{app:factorial}), and 56\% on 40 items for Gemma-2-2b, which often begins its answer with a newline. When we score the plain-text factorial from parsed generated answers, every change from base to aligned model keeps its direction, but OLMo-2's increase is no longer significant ($+0.07$, $p=0.21$, Appendix~\ref{app:factorial}).

\subsection{Pairing and tests}
\label{app:tests}
Comparisons between base and aligned models and between stages use only items that both models answered correctly without the cue, with paired differences per item, while the layer knockout scores each model on its own known items. Caving comparisons use exact McNemar tests, and susceptibility comparisons use a paired $t$ test with $n$ equal to the number of items. On the graded, arithmetic, and 70B sets, each item's $\Delta p_X$ is averaged over the four mid-strength wordings of the ladder (hedge, admission of ignorance, random guess, and plain opinion). Presentations of the same arithmetic problem are averaged into one unit, while the chain-of-thought comparison (Appendix~\ref{app:cot}) counts presentations.

\subsection{Attention measures}
\label{app:attn}
The reader attention rate is the share of items on which a designated reader's largest attention weight falls on a token that spells the cued option's letter. We average it over readers and measure it with eager attention at the final prompt position. The cued letter also appears outside the cue, in the list ``(A, B, C, D, \ldots)'' of the answer instruction (in 92\% to 95\% of items) and in the option labels. Letter-based measures therefore partly track the option the model is about to choose. On the graded items with no cue at all, the Tulu and OLMo readers' rate is 0.22 to 0.47 on items that the model answers with the cued letter, against 0.04 on items that it answers correctly. With a cue, it is 0.66 to 0.91 on caved items and 0.14 to 0.18 on resisted items. Several analyses use such measures. These are the stage-wise attention changes, the attention rates of the fine-tuning and role experiments, the attention-matching target, the checkpoint fingerprint, the item-level prediction, and the monitor. Their correlations with caving therefore overstate how much the heads read the cue. The causal claims rest on interventions that measure behavior directly. These are patching, weight copies, activation steering, the attention-pattern swap, and scaling of attention to the cue.

\subsection{Interventions}
\label{app:interventions}
Activation patching replaces a head's output at the answer position with its value from the paired uncued run. Its \emph{recovery} is the fraction of cue-induced flips whose answer returns to $T$. For interventions that also act on uncued prompts, \emph{preservation} is the fraction of uncued prompts that keep their correct answer. Random controls are head sets of the same size drawn with fixed seeds (1 to 30 sets per condition). In the cross-family set-level patches, a new set is drawn for every flip. Attention scaling adds $\log\lambda$ to the attention logits from each position after the cue to each cue token. It uses an additive mask in every layer and head unless a smaller set is stated. This multiplies the cue tokens' unnormalized attention weights by $\lambda$ before the softmax renormalizes each row. At $\lambda{=}0$ these links are removed, and every other link stays intact.

\section{The heads exist before alignment}
\label{app:circuit}
This section supports Section~\ref{sec:circuit}.

\subsection{Head discovery and the circuit in each family}
\label{app:replication}
\begin{table*}[t]
\centering\small
\setlength{\tabcolsep}{5pt}
\begin{tabular}{lllccc}
\toprule
family & example readers & writer hub & top-8 & top-16 & monitor AUROC \\
\midrule
gemma-2-2b & L18.H6, L14.H2, L13.H6 & L17.H3 & 0.80$^{*}$ & --- & 0.995 \\
gemma-2-9b & L20.H14, L24.H10, L26.H11 & (distributed) & 0.69 & 0.88 & 0.995 \\
llama-3.1-8b & L18.H28, L16.H19, L20.H14 & L17.H24 & 0.92 & 0.94 & 0.869 \\
qwen-2.5-7b & L22.H25, L21.H14, L22.H7 & L26.H26, L24.H27 & 0.85 & 0.96 & 0.911 \\
qwen-2.5-14b & (distributed) & (distributed) & 0.44 & 0.58 & --- \\
olmo-2-7b & L20.H7, L18.H19, L21.H18 & L18.H19$^{\dagger}$ & --- & --- & 0.954 \\
\bottomrule
\end{tabular}
\caption{Circuit summary for each family. Top-$k$ values are held-out recovery rates for patching the top $k$ heads (for Llama, two of the ten evaluation flips per cell of cue and dataset were also in the selection slice). $^{*}$The set of 7 circuit heads on suggested-answer flips that were not used for head selection (one random 7-head set gives 0.00). On the selection slice the set recovers 0.89 (bootstrap CI $[0.81, 0.96]$), and 30 random draws of the same size give a mean of 0.04 and a maximum of 0.29. The monitor head is selected on a labeled training half (Gemma-2-9b L30.H9, Qwen-7B L23.H11, OLMo-2 L20.H7, Tulu-SFT L20.H14), and Gemma-2-2b and Llama use fixed heads found by patching. $^{\dagger}$In OLMo-2 the strongest head both reads and writes (single-head recovery 0.63 at the SFT stage and 0.53 at the instruct stage). The main persuasion decision heads of \citet{sun2026persuaded} have the same indices as our hub heads, namely L17.H24 in Llama-3-8B-Instruct (the predecessor of the Llama-3.1 model we study) and L18.H19 in the same OLMo-2-1124-7B-Instruct checkpoint. Their Llama auxiliary decision heads include L15.H8 and L15.H11, two of our Llama writers.}
\label{tab:replication}
\end{table*}

For each model we collect its flips, as defined in \S\ref{sec:setup}. The four cross-family models pool all four cue types (Qwen-14B ranks heads on two and is evaluated on all four), and the Gemma-2-2b heads were selected on suggested-answer flips 12 to 31 and validated on flips 32 to 51. We patch every head in a fixed band of middle layers one at a time and rank the heads by recovery. Heads whose largest attention weight most often falls on the cued letter are designated readers (Appendix~\ref{app:attn}), and heads with high recovery whose attention is mostly elsewhere are designated writers. The reported recovery patches the top $k$ heads together ($k \in \{8, 16\}$, or the seven-head set for Gemma-2-2b) on flips whose indices differ from the discovery slice. Two caveats apply. In Llama, two of the ten evaluation flips per cell of cue and dataset were also in the selection slice. In the four cross-family models, a flip can also share its question with a discovery flip under another cue type. The discovery bands (layers 10 to 22 in Llama, 13 to 29 in Gemma-9B, 8 to 20 in Qwen-7B, 18 to 36 in Qwen-14B, and 13 to 18 in Gemma-2-2b) cover roughly 40 percent of each model's depth, except the older and narrower Gemma-2-2b band, and were fixed before any ranking. The rankings for Qwen-7B and Gemma-9B also pool extension scans (layers 14 to 26 and 24 to 36, on suggested-answer flips) that we added after inspecting their main-band rankings, and the Qwen-7B readers and writers in Table~\ref{tab:replication} come from its extension scan.

Table~\ref{tab:replication} summarizes the circuit in each family. Random head sets of the same size recover 0.02 to 0.05 on average for the top-8 patch and 0.03 to 0.08 for the top-16 patch, and the largest draw-series means are 0.08 and 0.11. For the Gemma-2-2b set, 30 fixed random sets reach at most 0.29 on the selection slice. What the families share is a small set of heads in middle and late layers. The head indices differ across families, and Qwen-14B is much more spread out. In plain text, the strongest of each base model's three readers ranks first or second in attention to the cued letter among all heads of a fixed layer band, and the two strongest of the three readers rank between 1 and 7. This is a descriptive attention measure, and we did not patch single heads in base models. Qwen-14B has no designated readers, so for it we use the three heads of its instruct model that attend most to the cue.

In Gemma-9B and Qwen-7B, the listed readers have near-zero recovery on their own, and so do two of the three Llama readers. Of the listed readers in these three models, only Llama's L18.H28 is among the top-8 patched heads, so we test the flow from readers to writers only in Gemma-2-2b. Patching the early readers (L13.H6, L13.H2, L14.H2) reverses 17 of 76 suggested-answer flips. With the writers (L15.H7, L16.H4, L17.H3) frozen at their biased values, it reverses 1. A random set of three heads from the circuit band (L13 to L18) reverses 10, and 3 with the writers frozen. The writers therefore also pass on the effects of other heads in the band, so the path through the writers is not specific to the readers.

\subsection{Leave-one-cue-out selection and single heads}
\label{app:loco}
To test whether the circuit is shared causally or is a union of cue-specific parts, we rank heads using three of the four cue types and patch the top eight together on the flips of the fourth. Recovery on the held-out cue is 0.90 (suggested answer), 0.98 (prior commitment), 0.81 (distractor fact), and 0.93 (wrong few-shot) in Llama, and 0.75, 0.60, 0.69, and 0.69 in Gemma-9B, against random means of 0.01 to 0.06 (20 random draws, 32 to 40 flips per cell). Llama's top two heads and Gemma-9B's top three are the same in every fold. The rule that we set in advance compared each cell with the pooled top-8 recovery on that cue. We did not store that value. Every cell is within 15 points of the recovery pooled over all four cues (0.92 in Llama, 0.69 in Gemma-9B). No single head is necessary. In four of five models, mean-ablating the best of 40 to 50 searched heads lowers caving by at most 0.01 more than ablating a random head. In OLMo-2-7B the best head, L18.H19, lowers caving on items still answered correctly from 0.58 to 0.48, but it also lowers uncued accuracy from 1.00 to 0.83.

\subsection{Weight copies and layer knockout}
\label{app:copies}
The weight copies take the query and output projections of the designated heads from one model and put them into the other. These heads are three readers and three writers, which in Gemma-2-2b leaves out L13.H2, the fourth reader of the seven-head set. We then rerun the full evaluation on flip items that include the discovery slice (400 flip items of Llama-3.1-8B-Instruct for Tulu-3-8B-SFT, and its own flips for Gemma-2-2b-it), with random six-head sets as the control. Separate arms copy the heads' shared key-value groups (Tulu) or one whole attention layer. Across the head and whole-layer arms, the aligned model with base weights caves at 0.81 to 0.82, compared with 0.80 without the copy (Tulu-3-SFT). For Gemma-2-2b-it the values are 0.95 to 0.99, compared with 1.00. The key-value arms give the same result. As a positive control, copying the SFT checkpoint's weights into the Llama base changes chat-format behavior (320 flip items). The MLPs of layers 0 and 1 alone raise caving from 0.21 to 0.57 and uncued accuracy from 0.03 to 0.22, and whole layers 0 to 11 raise them to 0.90 and 0.72 (the SFT checkpoint is at 0.80 and 0.74).

The layer knockout replaces one layer's attention output at the answer position with its uncued value, for every layer in turn. The most effective layer is the same in base and aligned weights in three of five models (layer 28 in Gemma-2-9B, 17 in Llama-3.1-8B, and 36 in Qwen-2.5-14B), and it differs in Gemma-2-2b (18 against 16) and Qwen-2.5-7B (19 against 24). On average, knocking out this layer removes 51\% (base) and 54\% (aligned) of caving (paired $t$-test over the five models, $p=0.82$, with wide intervals per model).

\subsection{Base and aligned models with format held fixed}
\label{app:factorial}
\begin{table*}[t]
\centering\small
\begin{tabular}{llrrrrr}
\toprule
family & format & $n$ & base & aligned & discordant & $p$ (McNemar) \\
\midrule
gemma-2-2b & plain text & 93 & 0.699 & 0.624 & 22/15 & $0.32$ \\
gemma-2-9b & plain text & 207 & 0.705 & 0.251 & 100/6 & $4\times10^{-23}$ \\
llama-3.1-8b & plain text & 173 & 0.751 & 0.584 & 36/7 & $9\times10^{-6}$ \\
qwen-2.5-7b & plain text & 254 & 0.508 & 0.193 & 87/7 & $1\times10^{-18}$ \\
qwen-2.5-14b & plain text & 294 & 0.248 & 0.078 & 55/5 & $1\times10^{-11}$ \\
olmo-2-7b & plain text & 142 & 0.331 & 0.430 & 14/28 & $0.04$ \\
\midrule
gemma-2-2b & chat & 37 & 0.000 & 0.622 & 0/23 & $2\times10^{-7}$ \\
gemma-2-9b & chat & 51 & 0.020 & 0.373 & 1/19 & $4\times10^{-5}$ \\
llama-3.1-8b & chat & 40 & 0.000 & 0.475 & 0/19 & $4\times10^{-6}$ \\
qwen-2.5-7b & chat & 207 & 0.459 & 0.145 & 76/11 & $4\times10^{-13}$ \\
qwen-2.5-14b & chat & 88 & 0.205 & 0.148 & 14/9 & $0.40$ \\
olmo-2-7b & chat & 31 & 0.194 & 0.774 & 2/20 & $1\times10^{-4}$ \\
\bottomrule
\end{tabular}
\caption{Caving rates for base and aligned models on matched items that both answered correctly without the cue, with format held fixed (Figure~\ref{fig:phenom}b). We exclude the 21 suggested-answer items, of 476 in all, whose cue the loader removed. In chat format, four base models rarely answer correctly (uncued accuracy 0.11 to 0.16 on these items), and for three of them the comparison reverses sign. The chat-format cells for these families are small precisely because the base rarely answers correctly in that format. Scored from parsed greedy generations, the plain-text rows keep their direction. Base and aligned caving are 0.65 and 0.58 for gemma-2-2b ($p=0.26$), 0.57 and 0.24 for gemma-2-9b, 0.81 and 0.58 for llama-3.1-8b, 0.35 and 0.15 for qwen-2.5-7b, and 0.25 and 0.08 for qwen-2.5-14b, each with $p<10^{-7}$ except gemma-2-2b, and 0.37 and 0.44 for olmo-2-7b ($p=0.21$). When scored from generations, these four base models answer 3\% to 19\% of uncued prompts correctly in chat format.}
\label{tab:factorial}
\end{table*}

Table~\ref{tab:factorial} gives the caving rates behind Figure~\ref{fig:phenom}b.

\subsection{Beyond multiple choice}
\label{app:beyondmcq}
\label{app:freeform}
On SycophancyEval answer sycophancy (open TriviaQA questions with a stated user belief, scored by the log-probability margin between the correct and the incorrect answer after ``The answer is''), 70.6\% of Gemma-2-2b's known answers flip (154 of 218), and 25.5\% of Llama-3.1-8B's (56 of 220). Patching the circuit heads recovers 15.6\% of Gemma's flips, against 1.3 to 7.1\% for five random sets of the same size. In Llama the patch reverses only 4 of 56 flips (7.1\%), one more than the best random set (random sets reverse 0 to 5.4\%). The circuit's writers alone reverse 10 (17.9\%). The effect also appears when the cue names no letter. When the prompt asks for the text of the option and the cue states the wrong option's content, the patch still reverses 0.20 of flips in Gemma (random mean 0.037) and 0.15 in Llama (random mean 0.003). In Gemma, masking the six tokens that the writer hub L17.H3 attends to most restores correct free-form answers as well as masking the exact tokens of the suggested wrong answer does (0.880, compared with 0.870). A head chosen without labels to locate the cue (L16.H7) restores 0.65. Uncued accuracy on these 108 known items falls only to 0.84 (0.87 when masking a random span), while deleting the same tokens leaves 0.19.

\section{Generic assistant tuning activates the heads in chat format}
\label{app:activation}
This section supports Section~\ref{sec:gate}.

\subsection{Short fine-tuning runs}
\label{app:finetune}
The runs train rank-16 LoRA adapters ($\alpha=32$, dropout 0, on all attention and MLP projections) on the Llama-3.1-8B base with cue-free generic or code subsets of the Tulu-3 SFT mixture. They use AdamW at learning rate $2\times10^{-4}$ with no weight decay, a cosine schedule with 3\% warmup, batch size 8, gradient clipping at 1.0, sequences of up to 1024 tokens, and loss on the assistant response only, for two epochs. At $N=200$ this is 50 optimization steps. The comparison with the full SFT checkpoint therefore shows how little data is enough to reach SFT-level caving, and it says nothing about efficiency at matched compute. Draws differ in the training sample. Evaluation uses the chat template, on 400 flip items of Llama-3.1-8B-Instruct (25 per cue type and dataset), and for Gemma on Gemma-2-2b-it's own flips. The adapters were merged into the bf16 base weights before evaluation and were not kept. Merging rounds the update to bf16, and we did not measure how much this changes the results.

With 200 generic examples, the attention rate averaged over the three readers is 0.65 to 0.76 over three draws, against 0.55 for the full SFT checkpoint, and caving is 0.86 to 0.94 against 0.80. With 200 code examples (five draws), caving is 0.81 to 0.91, with a lower attention rate. Figure~\ref{fig:evidence}b shows single draws at 20, 50, 1,000, and 5,000 examples and the mean and range of three draws at 200. We repeat this in Gemma-2-2b with the same data and procedure. At 200 examples, the rate of the main reader (L18.H6) goes from 0.00 to 0.79, compared with 0.90 for the tuned checkpoint. Caving reaches 0.94, compared with 1.00 for the tuned checkpoint. These items are the tuned checkpoint's own flips. The model tuned on 200 examples answers only 0.29 of them correctly without the cue. On the unselected fixed set of the role control (Appendix~\ref{app:rolectl}), susceptibility after 200 examples matches the tuned checkpoint in both families (0.47 in Llama and 0.45 in Gemma). To test whether the same heads carry the new caving, we patch the top eight heads of Llama-3.1-8B-Instruct (Appendix~\ref{app:replication}) in each tuned model, replacing their answer-position outputs with their uncued values. The item pool has 20 to 30 flip items of Llama-3.1-8B-Instruct per cue type and dataset, including the 12 items per cell used to select these heads, and we patch on the items in this pool that the tuned model flips. The eight heads undo 0.54, 0.69, and 0.73 of the flips in the three generic models (184, 156, and 147 flips) and 0.37 to 0.58 in the five code models (75 to 169 flips). Five random sets of eight heads from layers 13 to 22 undo at most 0.03 in the generic models and 0.04 in the code models. The three designated readers alone undo at most 0.04.

\subsection{Role control}
\label{app:rolectl}
\begin{table*}[t]
\centering
\small
\setlength{\tabcolsep}{3.5pt}
\begin{tabular}{lccccc}
\toprule
Condition & Uncued acc. & Caving & Susceptibility & Flips (known) & Reader attention \\
\midrule
Base, plain text & 0.48 & 0.83 & $+0.31$ & 0.77 \; (183) & 0.40 \\
Base, chat, no prefix & 0.03 & 0.23 & $+0.06$ & 0.00 \; (11) & 0.00 \\
Base, chat, 8 role turns & 0.25 & 0.62 & $+0.14$ & 0.46 \; (97) & 0.11 \\
Base, chat, same text without roles & 0.14 & 0.33 & $+0.04$ & 0.00 \; (53) & 0.02 \\
Tuned (SFT), chat & 0.56 & 0.68 & $+0.47$ & 0.63 \; (216) & 0.49 \\
\bottomrule
\end{tabular}
\caption{\textbf{Eight generic dialogue turns lose their activating effect when the role structure is removed (unselected items).} The prefix without roles stays near the inactive level in susceptibility, known-item flips, and reader attention. Fixed-set susceptibility does not condition on correctness. Flip rates are over items answered correctly without the cue, with counts in parentheses.}
\label{tab:rolectl}
\end{table*}

We evaluate the Llama-3.1-8B base model in the five conditions of Table~\ref{tab:rolectl}, with its SFT checkpoint as a reference. The 400 unselected items were fixed before any model was run. They are 100 questions, each under all four cue types, taken as the first 25 per cue type and dataset, and 383 remain after we exclude the 17 cue-less items. The roleless condition joins the same eight generic question-answer exchanges as `Q:'/`A:' pairs inside the first user message, with the chat role markers removed. Susceptibility is computed on the full fixed set, so it does not depend on which items each condition answers correctly.

The original run on flip items had a rule, set in advance, that separated two outcomes. Either the role structure matters, or generic content is enough. For this rerun on unselected items, we set fixed-set caving and susceptibility as the primary contrasts in advance. The data match the first outcome. The result also holds on matched items. On the 22 items answered correctly without the cue in the plain-text, role-turn, and roleless conditions, flips are 0.86, 0.45, and 0.00. In a larger accuracy-matched comparison, not specified in advance, role turns flip 19 of 41 items (11 questions) and the roleless prefix none (exact McNemar $p=4\times10^{-6}$), and role turns flip more items in all 10 questions where the two prefixes differ (sign test $p=0.002$). We also tried eight role turns in which a user states a wrong fact and the assistant agrees with it or corrects it. On the flip items, the base then almost never gives the correct or the cued answer. With the agreeing and correcting turns, caving is 0.01 and 0.00 and uncued accuracy is 0.05 and 0.03. With the generic turns, caving is 0.69 and uncued accuracy is 0.32. This test therefore cannot show whether role structure alone is enough. Because the generic role turns also raise uncued accuracy (0.03 to 0.25), our claim is only that these generic turns produce caving when written as dialogue. In Gemma-2-2b the role contrast replicates, but more weakly. On 42 accuracy-matched items (11 questions), role turns flip 7 and the roleless prefix none (exact McNemar $p=0.016$, sign test $p=0.03$). On these items, however, susceptibility is near zero under both ($+0.004$ and $+0.014$). On the full fixed set it is $+0.04$ with role turns and $+0.01$ without roles. Only 12 items are known in all three conditions. Plain text flips 8 of them, and the role turns and the roleless prefix flip none. This is too few to settle the role contrast on matched items.

\subsection{Activation directions}
\label{app:directions}
The vectors are differences in the residual stream at the answer position, at one early layer $L \in \{10, 12, 14\}$ of Llama-3.1-8B, over prompts that contain no cue. $v_{sft}$ is the mean residual under the SFT checkpoint minus the mean under the base, and $v_{icl}$ is the mean residual of the base with eight generic dialogue turns added before the prompt, minus the base without them. Steering adds $\alpha v$ at that layer and position. The original run tried $\alpha \in \{1, 2, 3\}$. For $v_{sft}$ and $v_{icl}$, a larger $\alpha$ never raised caving, and $\alpha=3$ lowered accuracy at every layer. We therefore report $\alpha=1$, the only value that the item-disjoint run uses. The control adds one seeded Gaussian direction whose expected norm equals that of $v_{sft}$. Figure~\ref{fig:switch} reports the item-disjoint run that we specified in advance. It extracts the vectors from a random half of the items and evaluates steering only on the other 120 items (45 of which share their uncued prompt with an extraction item under another cue). In this run, $v_{sft}$ produces caving of 0.87 (layer 12) and 0.88 (layer 14), and $v_{icl}$ produces 0.53 and 0.62, against 0.18 and 0.17 for the random direction (the base model is at 0.18). The cosine between $v_{sft}$ and $v_{icl}$ is 0.64 and 0.72. The SFT shift on generic prompts that are not multiple-choice has a similar cosine with $v_{sft}$ (0.70 and 0.68), yet it produces little caving (0.28). The cosine alone therefore tells us little. The original run on all 240 items gave 0.87 for $v_{sft}$ and 0.65 for $v_{icl}$ at their best layers, against 0.21 to 0.22 for a random direction.

\subsection{Patching after in-context activation}
\label{app:iclpatch}
We place eight generic dialogue turns, or three multiple-choice examples, before the prompt of the Llama-3.1-8B base in the Tulu-3 chat template, on up to 40 flip items of Llama-3.1-8B-Instruct per cue type and dataset, which include the 12 items per cell used to select the heads patched below. The flips caused by the context are the items that the base with this context answers correctly without the cue and with $X$ under it (122 with the dialogue turns and 186 with the examples). We patch the answer-position outputs of the three readers (L18.H28, L16.H19, L20.H14) and three writers (L17.H24, L15.H8, L15.H11) with their uncued values, and compare with 10 random sets of six heads from layers 13 to 20 outside these heads. The six heads undo 53\% and 60\% of these flips, the writers alone 48\% and 54\%, the readers alone 2\% and 3\%, and the random sets 2\% and 1\% on average (at most 7\%).

\section{The final RL stage strengthens the effect of cues}
\label{app:stage}
This section supports Section~\ref{sec:gain}.

\subsection{Stage tests}
\label{app:grid}
\begin{table}[t]
\centering\small
\setlength{\tabcolsep}{4pt}
\begin{tabular}{lrrr}
\toprule
test & $n$ & $\Delta$susc & $t$ \\
\midrule
graded 8B (Tulu) & 232 & $+0.075$ & $+14.6$ \\
graded 7B (OLMo) & 212 & $+0.074$ & $+15.8$ \\
graded 13B (OLMo) & 244 & $+0.015$ & $+4.3$ \\
graded 32B (OLMo) & 305 & $+0.031$ & $+8.9$ \\
published 8B (Tulu) & 241 & $+0.050$ & $+5.5$ \\
published 7B (OLMo) & 243 & $+0.053$ & $+7.7$ \\
published 13B (OLMo) & 253 & $+0.009$ & $+2.0$ \\
published 32B (OLMo) & 304 & $+0.036$ & $+5.5$ \\
held-out cues 8B (Tulu) & 125 & $+0.071$ & $+4.9$ \\
held-out cues 7B (OLMo) & 121 & $+0.036$ & $+4.8$ \\
arithmetic 8B (Tulu) & 57 & $+0.045$ & $+7.1$ \\
arithmetic 7B (OLMo) & 28 & $+0.116$ & $+8.6$ \\
graded 70B (Tulu) & 123 & $-0.001$ & $-1.0$ \\
\bottomrule
\end{tabular}
\caption{The 13 stage tests. Each row gives the paired change in susceptibility ($\Delta p_X$) from DPO to the final stage (RLVR for Tulu, Instruct for OLMo) on items that both stages answer correctly without the cue. Graded, arithmetic, and 70B rows average the four mid-strength cue wordings, and the arithmetic $n$ counts unique problems. Twelve of the 13 changes are positive, and 11 are significant after Bonferroni correction. No change in uncued confidence survives the correction. Also counting fused letter tokens such as ``(B'' shrinks the Tulu-8B published and held-out differences by a quarter to a third, but both stay significant after correction (Appendix~\ref{app:readout}).}
\label{tab:grid}
\end{table}

\begin{table*}[t]
\centering\small
\begin{tabular}{lrrrr}
\toprule
$\Delta$susc ($t$) & Tulu 8B & OLMo 7B & OLMo 13B & OLMo 32B \\
\midrule
suggested answer & $+.009$ (0.6) & $+.042$ (5.0) & $+.016$ (1.4) & $+.026$ (2.5) \\
prior commitment & $+.021$ (1.7) & $+.030$ (5.2) & $+.015$ (2.2) & $+.049$ (4.1) \\
distractor fact & $+.032$ (3.7) & $+.078$ (7.1) & $-.002$ ($-0.5$) & $+.055$ (4.5) \\
wrong few-shot & $+.132$ (5.0) & $+.060$ (2.7) & $+.010$ (0.7) & $+.013$ (0.8) \\
held-out: distractor argument & $+.069$ (3.2) & $+.046$ (6.8) & --- & --- \\
held-out: spurious few-shot & $+.073$ (3.7) & $+.026$ (2.0) & --- & --- \\
\bottomrule
\end{tabular}
\caption{Change in susceptibility from DPO to the RL stage for each cue type (paired, matched items, $n=52$ to $80$ per cell). Nineteen of the 20 cells are positive. The exception is OLMo-13B distractor fact ($-0.002$), and OLMo-13B shows smaller changes than the other models overall. Which cue types rise most differs by recipe.}
\label{tab:percue}
\end{table*}

The stage comparisons run the same items through each public checkpoint of a pipeline in its released chat template. These templates are identical within each pipeline, except that OLMo-2-32B's DPO template adds a BOS token. The 70B runs use four A100 GPUs, the 32B runs two, and all other runs one. Table~\ref{tab:grid} lists all 13 stage tests, and Table~\ref{tab:percue} splits the effect by cue type. The one positive test that is not significant after correction, OLMo-13B on the published cues, has $p=0.048$ before correction. No change in uncued confidence survives Bonferroni correction, and before correction confidence falls in three tests and rises in one (Tulu published, $+0.011$). The 70B RLVR update changes 1.8\% of the stored bf16 parameters. Its relative weight change is $1.6\times10^{-5}$, 45 times smaller than at the 8B RL stage. Meta's instruct model, tuned separately from the same Llama-3.1-8B base, has lower susceptibility than Tulu's RLVR stage on matched items ($-0.154$, $t=-9.0$, $n=217$ on the graded set, and $-0.161$, $t=-6.6$, $n=220$ on the published cues). Zephyr, whose pipeline has no RL stage, shows a positive but not significant change from SFT to DPO on the graded set ($+0.049$, $t=1.6$, $n=100$), and a change of $+0.054$ ($t=4.6$) from base to SFT.

\subsection{Attention matching}
\label{app:gainmatch}
\begin{table*}[t]
\centering
\small
\setlength{\tabcolsep}{4.5pt}
\begin{tabular}{llccc}
\toprule
Pipeline & Direction & Chosen $\lambda$ & Held-out attention rate (matched / target) & $M$ (95\% CI) \\
\midrule
Tulu DPO$\to$RLVR & RLVR down & 0.85 & 0.434 / 0.432 & 0.51 (0.30, 1.00) \\
 & DPO up & 1.3 & 0.468 / 0.477 & 0.80 (0.50, 1.52) \\
OLMo DPO$\to$Instruct & Instruct down & 0.7 & 0.631 / 0.674 & 1.39 (1.11, 1.72) \\
 & DPO up & 1.7 & 0.737 / 0.714 & 1.23 (0.89, 1.62) \\
\bottomrule
\end{tabular}
\caption{\textbf{Attention matching transfers at least half of the stage difference in susceptibility.} The scale $\lambda$ is chosen on calibration items from attention alone. $M$ is the transferred fraction of the susceptibility difference between DPO and the final stage on held-out items. The held-out attention rate averages all 44 held-out items.}
\label{tab:gainmatch}
\end{table*}

Attention matching gives one stage the other stage's attention to the cue and asks whether it also takes on the other stage's susceptibility. We split suggested-answer items in half by question, into 44 calibration and 44 held-out items per pipeline, and average susceptibility over the four mid-strength wordings. On the calibration half, we sweep $\lambda$ over $\{1, 0.85, 0.7, 0.55, 0.4, 0.25\}$ downward and $\{1, 1.3, 1.7, 2.2, 3, 4\}$ upward and choose the grid point whose reader attention rate is closest to the other stage's natural level (Tulu DPO 0.509 and RLVR 0.566, OLMo DPO 0.650 and Instruct 0.701). No behavioral quantity enters this choice, and on the calibration half all four chosen points land within 0.015 of their targets. On the held-out items that both natural stages answered correctly without the cue ($n=24$ Tulu, $n=26$ OLMo), we compute $M_\downarrow = 1 - (S[\text{RL}@\lambda_\downarrow]-S[\text{DPO}])/(S[\text{RL}]-S[\text{DPO}])$ and $M_\uparrow = (S[\text{DPO}@\lambda_\uparrow]-S[\text{DPO}])/(S[\text{RL}]-S[\text{DPO}])$, with percentile bootstrap CIs over items (1000 draws). Following our rules (Appendix~\ref{app:prereg}), transfer counts as substantial at $M \ge 0.6$, partial at 0.3 to 0.6, and null below 0.3 or with a CI spanning zero. Three cells are substantial, and the Tulu downward cell is partial (Table~\ref{tab:gainmatch}). The criterion, set in advance, that all four cells reach 0.6 is therefore not met, and the main text states the claim at the level that the weakest cell supports. The OLMo estimates exceed 1 because matching downward pushes the Instruct stage's susceptibility below DPO's natural level, and DPO scaled up rises above the Instruct stage's level. The natural stage differences are small ($0.06$ in both pipelines), so $M$ is a ratio with a small denominator and wide intervals.

\subsection{Where the difference transfers}
\label{app:gainlocal}
\begin{table*}[t]
\centering
\small
\setlength{\tabcolsep}{4pt}
\begin{tabular}{lcccc}
\toprule
Layer set & Tulu $M_\downarrow$ & Tulu $M_\uparrow$ & OLMo $M_\downarrow$ & OLMo $M_\uparrow$ \\
\midrule
reader heads (3) & $-0.06$ & $-0.07$ & $0.40$ & $0.42$ \\
readers + writers (5) & $0.09$ & $0.12$ & --- & --- \\
reader layers (all heads) & $0.06$ & $0.03$ & $0.69$ & $0.69$ \\
middle band (6 layers) & $\mathbf{0.97}$ {\scriptsize(0.73, 1.35)} & $\mathbf{0.99}$ {\scriptsize(0.69, 1.45)} & $\mathbf{0.75}$ {\scriptsize(0.61, 0.93)} & $\mathbf{0.92}$ {\scriptsize(0.53, 1.37)} \\
low band (6 layers) & n.m. & n.m. & $0.09$ & $\lambda{=}1$ \\
high band (6 layers) & n.m. & n.m. & $\lambda{=}1$ & $\lambda{=}1$ \\
random 3 layers & $0.07^{\dagger}$ & $0.07$ & $0.06$ & $0.17$ \\
random 6 layers (a) & n.m. & n.m. & $-0.01$ & $\lambda{=}1$ \\
random 6 layers (b) & n.m. & n.m. & $0.03$ & $\lambda{=}1$ \\
\midrule
network-wide (reference) & $0.51$ & $0.80$ & $1.39$ & $1.23$ \\
\bottomrule
\end{tabular}
\caption{\textbf{Most of the stage difference transfers through the six layers directly below the lowest reader layer}, the middle band (Tulu L10--15, OLMo L12--17). The table gives transferred fractions, with 95\% CIs for the middle band. ``---'' means not run. ``n.m.'' means that no grid point brings the reader attention rate within 0.05 of the target, so the layer set receives no transfer claim. The natural attention gaps between the stages are 0.052 (Tulu) and 0.047 (OLMo), about the size of this tolerance. On the calibration questions, the middle band, reader layers, reader heads, and circuit heads close at least 66\% of the gap (some overshoot it). Every control set closes 43\% or less, so the controls give no matched comparison. $\lambda{=}1$ means that the nearest grid point is the natural condition, because the layer set moves attention so little. That arm is then the natural run. $\dagger$ marks a calibration miss of 0.052, just past the threshold. The random three-layer sets exclude only reader layers, and Tulu's set (L4, L5, L10) includes the middle-band layer L10. The network-wide row repeats Table~\ref{tab:gainmatch} (a separate run on 44 held-out questions per pipeline). It is the reference for the 0.6 criterion that we set in advance. The other rows come from the layer-set sweep, and the text reports a larger separate run at the reader heads.}
\label{tab:gainlocal}
\end{table*}

A layer-set sweep, specified in advance, repeats this procedure with attention to the cue scaled within one layer set at a time (121 calibration and 121 held-out questions per pipeline, $n=57$ and $54$ matched items, Table~\ref{tab:gainlocal}). The random six-layer sets exclude the middle-band, reader, and writer layers. In Tulu, three kinds of layer set cannot bring the reader attention rate to its target. These are the band below the middle band, the band above it, and the random six-layer sets. The band above comes after every reader layer, so by construction it cannot change the readers' attention. The natural attention gaps (0.052 in Tulu, 0.047 in OLMo) are about the size of the 0.05 tolerance. Control sets that reach the target within this tolerance therefore still close at most 43\% of the gap, so for specificity we rely on the pattern swap (Appendix~\ref{app:swap}). In Tulu, scaling at the reader layers steers the readers' attention as precisely as scaling at the middle band (calibration misses at or below 0.016). Yet it transfers almost none of the behavior. In Tulu, scaling at the reader layers is therefore not enough to move behavior, and scaling the middle band, the six layers below the reader layers, moves most of it. This parallels the finding of \citet{sun2026persuaded} that a shallow band of layers builds the feature that their decision heads use. For OLMo in the downward direction, the middle band reaches only 0.54 of the network-wide $M$, short of the 0.6 that we set in advance. OLMo's three reader layers as a whole transfer $M = 0.69$. A separate, larger run scales attention to the cue at the three reader heads only, in the final query row (184 calibration and 184 held-out questions). In OLMo it transfers $M_\downarrow = 0.36$ (CI 0.29 to 0.44) and $M_\uparrow = 0.37$ (0.30 to 0.45), and random non-reader heads in the same layers transfer $-0.01$ to $0.15$. In Tulu it transfers nothing ($M = -0.04$ and $-0.09$), even though the readers' attention moves as calibrated. Scaling at the reader heads therefore moves part of the difference in OLMo and moves no behavior toward the donor in Tulu.

\subsection{Attention-pattern swap}
\label{app:swap}
The swap runs identical prompts through both checkpoints and records each checkpoint's attention patterns after the softmax. It then reruns the other checkpoint with the patterns of the chosen layers fully replaced (all heads, all positions). There is no scale parameter. A checkpoint patched with its own patterns reproduces its natural logits to four decimal places on a check prompt. On the held-out questions of the layer-set sweep ($n=57$ and $54$), swapping the middle band transfers $M_\downarrow$ and $M_\uparrow$ of $0.72$ and $0.73$ in Tulu and $0.61$ and $0.62$ in OLMo, and all CIs exclude zero. Uncued answers stay correct under every swap of the middle band, the writer layer, or both (preservation 1.00), and the clean control cells keep 0.98 to 1.00.

The controls ran in two waves. In the first wave, ten layer sets contain no middle-band, reader, or writer layer, and they transfer between $-0.29$ and $+0.14$. Two Tulu random sets that contain the writer layer L17 transfer $0.63$ to $0.74$, so in Tulu the first wave failed the specificity rule that we set in advance. A second wave, specified in advance, then tested the writer layer separately from the middle band, a split that we found after the first wave. In Tulu, the writer layer alone transfers $0.68$ and $0.64$, and the writer layer together with the middle band transfers $0.88$ and $0.89$. Twenty new random six-layer sets per pipeline, drawn without the middle-band, reader, and writer layers, transfer less than $0.3$ in all eighty cells. OLMo has no separate writer layer, because its writer hub L18.H19 both reads and writes inside a reader layer. The difference is therefore largely transferred by the circuit region in the middle layers, and by none of the layer sets that we tested outside it. We found the writer-layer split on these same items, however. Swapping only the three reader layers moves behavior away from the donor in Tulu ($-0.29$ and $-0.27$) and partly toward it in OLMo ($+0.53$ and $+0.52$). In a first run, a generator error made six of the eight random layer sets overlap the middle band or the reader layers. We exclude all eight sets from the null result, and the released results keep them.

We also swap the middle band on the two held-out cue types. The stage difference is measurable in three of the four cells of pipeline and cue, and in the fourth (OLMo, spurious few-shot) its CI spans zero. In the three measurable cells, the swap transfers $M$ of $0.48$ to $0.78$, and every CI excludes zero. Of the 48 control cells of eight clean random six-layer sets per pipeline, 45 transfer less than $0.3$ in magnitude, and the exceptions, all in Tulu, reach $0.47$ (the control that failed the first run), $0.34$, and $-0.30$. A first run with two controls per cell had specified in advance that both stay below $0.3$, and one Tulu control reached $0.47$. The criterion for eight controls (at least 90\% of control cells below $0.3$) was set after that run and before the run with eight controls.

\subsection{Query, key, and value states}
\label{app:qkv}
Swapping the donor's query and key states together reproduces the pattern swap exactly, as a consistency check. The query side alone carries most of the difference in OLMo ($0.57$ and $0.54$) and the largest single share in Tulu ($0.50$ and $0.46$), and the key side carries less ($0.45$ and $0.35$ in Tulu, $0.22$ and $0.24$ in OLMo). Donor values under the receiving model's own patterns carry part of the difference in Tulu ($0.38$ and $0.33$) and less than 0.3 in OLMo ($0.21$ and $0.28$), below the partial band. Every CI excludes zero. These are tests of causal sufficiency, so the shares need not add up. The expectation set in advance that donor values fall in the null band (below 0.3) is not met in Tulu. We also test a single fixed query shift per middle-band layer. The shift is the mean difference between DPO and RL in the final-position query on the calibration half. It reproduces the query-side swap in Tulu ($0.51$ and $0.50$). In OLMo it reproduces only 40\% to 44\% of the query-side swap ($0.23$ and $0.24$), which misses the threshold of $0.3$ that we set in advance. Random shifts of the same norm transfer $-0.05$ to $0.00$.

\subsection{The middle band in the pretrained base}
\label{app:baseband}
In the pretrained base models in plain text, we scale the middle band's attention to the cue over $\lambda \in \{0, 0.25, 0.5, 1, 2, 4\}$. We measure susceptibility on the items that each base answers correctly without the cue (51 in Tulu, 50 in OLMo). In the Tulu base (Llama-3.1-8B) it is 0.05, 0.11, 0.14, 0.18, 0.21, and 0.23, rising at every step. In the OLMo base it is 0.02, 0.16, 0.25, 0.32, 0.34, and 0.32, dipping at 4. The rule that we set in advance required a middle-band response that rises at every step and is larger than the responses of the low and high bands. This rule holds in Tulu but not in OLMo. The high band is flat in OLMo and moves Tulu only slightly (0.17 to 0.20), and the low band is flat in Tulu while moving in the opposite direction in OLMo (0.35 down to 0.22). Scaling at the reader heads barely moves susceptibility in Tulu (0.19 to 0.17). In OLMo it also moves susceptibility (0.23 to 0.37), which fits the partial transfer through OLMo's reader heads (Appendix~\ref{app:gainlocal}).

\subsection{Reader attention across stages}
\label{app:rank}
On a subset of the graded items ($n=121$ Tulu, $114$ OLMo), the readers' attention rate to the cued letter rises from DPO to the RL stage (Tulu $+0.052$, $t=5.5$, and OLMo $+0.064$, $t=6.5$). On the published four-cue items, the rate rises in OLMo ($t=2.5$). In Tulu the rise is not significant ($t=1.1$), which misses the criterion that we set in advance (a positive change with $|t|>2$). Tulu's prior-commitment cell falls ($-0.058$, $t=-3.3$), while the susceptibility for that cue does not. We also measure the attention margin, the readers' attention mass on the cued option's letter minus their mass on the correct option's letter. It rises in both pipelines ($t=16.1$ in Tulu and $13.6$ in OLMo, on 150 and 139 items). The readers' total attention mass on the cue text rises in Tulu ($+0.024$, $t=15.3$) and falls in OLMo ($-0.009$, $t=-5.3$). In OLMo, however, the letter-based rate and caving both rise ($+0.064$ and $+0.057$). The letter-based measures partly track the chosen answer (Appendix~\ref{app:attn}), so this mismatch may partly reflect attention to the chosen answer.

\section{Confidence and reasoning}
\label{app:conf}
This section supports Section~\ref{sec:law}.

\subsection{Confidence directions}
\label{app:confdir}
\begin{figure*}[t]
\centering
\includegraphics[width=0.85\textwidth]{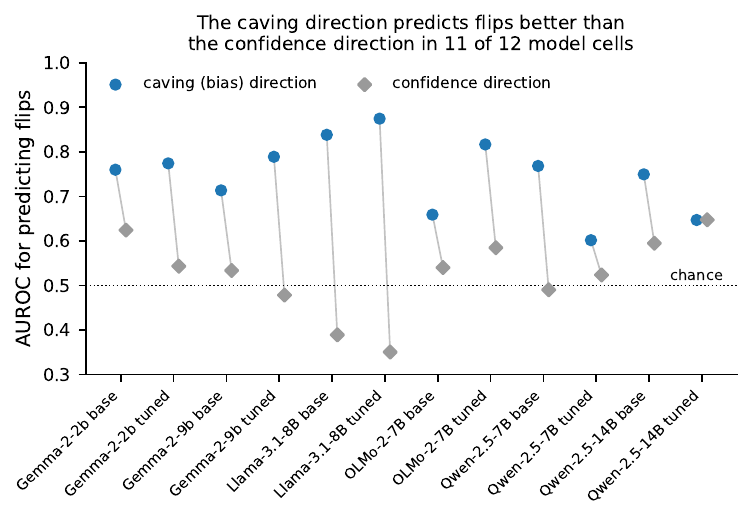}
\caption{\textbf{The caving direction predicts flips better than the confidence direction in 11 of 12 model cells.} Held-out AUROC, averaged over probed layers. The confidence direction is scored so that lower confidence predicts caving, as the confidence account expects, and it is ahead only in Qwen-2.5-14B tuned, by 0.0004.}
\label{fig:conf}
\end{figure*}

\begin{figure*}[t]
\centering
\includegraphics[width=0.6\textwidth]{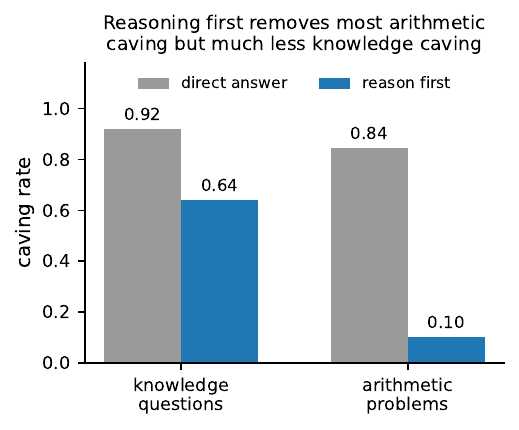}
\caption{\textbf{Reasoning first removes most caving on arithmetic but much less on knowledge questions.} Tulu-3-8B, pooled over DPO and RLVR, on items answered correctly in both answering modes, with identical cue wording (111 knowledge and 129 arithmetic presentations).}
\label{fig:cot}
\end{figure*}

The caving and confidence directions are differences in the residual stream at the answer position, probed at relative depths $\{0.35, 0.5, 0.65, 0.8\}$. The caving direction contrasts flipped and resisted items on cued prompts, and the confidence direction contrasts high-confidence and low-confidence items on uncued prompts, so it is built without any cue. Both are built on a training half that shares no questions with the evaluation half and are compared with random directions of the same norm. Steering adds $\alpha d$ at relative depths $\{0.4, 0.55, 0.7\}$ with $\alpha \in \{-8, -4, -2, -1, 0, 1, 2, 4\}$, with random and label-shuffled controls at the middle depth. A layer counts as usable when, at the values of $\alpha$ that keep uncued accuracy at 0.85 or more, the confidence direction moves uncued confidence by at least 0.03, with at least three such values.

In all 12 model cells (6 models, base and tuned), the caving direction predicts held-out flips (AUROC 0.60 to 0.87). The confidence direction barely does (Figure~\ref{fig:conf}, mean AUROC 0.53). It is below chance in four cells, most clearly in both Llama cells (0.39 and 0.35). Uncued confidence itself has a mean AUROC of 0.69. The two directions are only weakly related (mean cosine $-0.17$). Measured as distance from chance, the caving direction predicts flips better in 11 of 12 cells, and the confidence direction predicts uncued confidence better in 9 of 12. Steering along the confidence direction moves measured confidence by up to 0.41 (Gemma-2-9b-it, layer 29), and caving in that cell moves by at most 0.03. Across the four families, the slope of caving on confidence is negative in 7 and positive in 6 of the 13 usable cells. In some cells, steering moves caving. In Gemma-2-9b base at layer 23, raising confidence from 0.21 to 0.33 lowers caving from 0.70 to 0.55. Random and label-shuffled controls were run at the middle depth, and in 3 of the 4 usable middle-depth cells the confidence direction moves caving more than either control. The controls, however, also move confidence less (range over the usable values of $\alpha$, 0.01 to 0.12, compared with 0.11 to 0.23). Per unit of confidence moved, a control therefore changes caving at least as much in 3 of these 4 cells by the slope of caving on confidence, and in all 4 by the ratio of the two ranges.

\subsection{Reasoning before answering}
\label{app:cot}
We use identical cue wording (``I think the answer is X.'') and items answered correctly in both answering modes. On arithmetic, reasoning before answering cuts Tulu's caving from 0.79 to 0.19 at DPO and from 0.90 to 0.01 at RLVR. On knowledge questions it cuts caving only from 0.88 to 0.63 and from 0.96 to 0.65 (Figure~\ref{fig:cot} pools the stages). Pooled over stages, the interaction is $+0.46$ (unpaired $z=+7.0$). Its unit is one presentation of an item at a stage, with 111 knowledge and 129 arithmetic presentations. Counting each distinct arithmetic prompt once per stage gives $+0.55$ ($z=+7.4$). An automatic parser reads the final answers and misreads some chains that are cut off or that correct themselves, and we report the parsed values. This pattern is consistent with the idea that deriving an answer restores the margin more than reasoning about a knowledge question does, and it separates two kinds of question that \citet{feng2026goodarguments} pool as objective tasks. A keyword search flags 2 of the 109 caved knowledge chains in the Tulu-3 and OLMo-2 checkpoints, and when we read these two, neither refers to the cue. Nine of the stored traces are cut off at 1,200 characters. Reasoning models state such hints more often than non-reasoning models, though still in a minority of cases \citep{chen2025reasoning}.

\section{Attention scaling and two-turn retraction}
\label{app:probe}
\begin{figure*}[t]
\centering
\includegraphics[width=\textwidth]{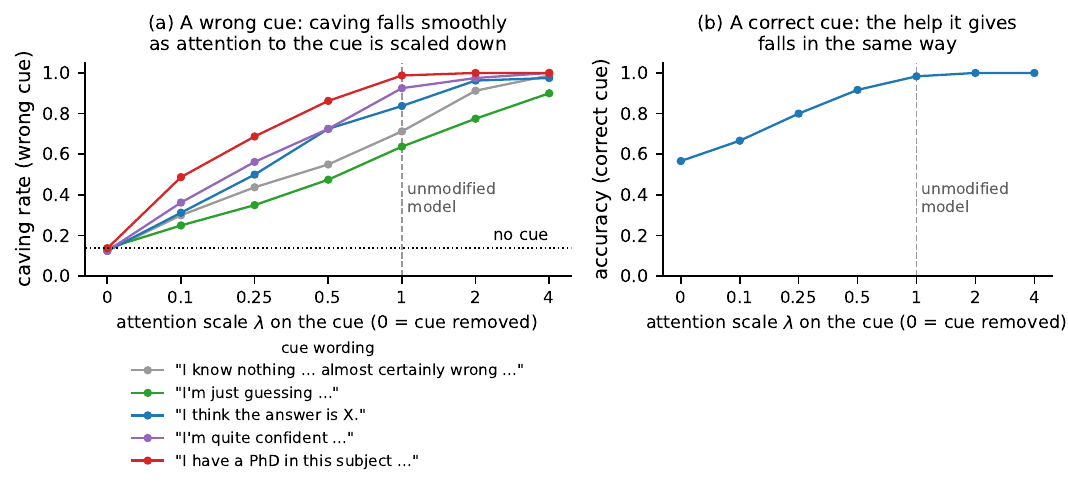}
\caption{\textbf{Scaling attention to the cue moves caving smoothly and also removes the help of a correct cue.} Gemma-2-2b-it, with attention to the cue scaled by $\lambda$ in every layer. \textbf{(a)}~Caving as the rate of answering $X$ on 80 unselected wrong-cue MMLU items, including items that the model does not know. \textbf{(b)}~Accuracy on 60 correct-cue items with the plain opinion cue.}
\label{fig:dialcurves}
\end{figure*}

This section supports Section~\ref{sec:dial}.

\subsection{The attention dial}
\label{app:dial}
Figure~\ref{fig:dialcurves} scales attention to the cue by $\lambda$ in every layer of Gemma-2-2b-it, on 80 unselected wrong-cue MMLU items and 60 correct-cue items. For the plain opinion cue, $\lambda=0.1$ removes 75\% of the caving that the cue adds, measured against the no-cue rate (0.84 at $\lambda=1$, 0.31 at $\lambda=0.1$, and 0.14 without a cue). It also removes 76\% of the accuracy that a correct cue adds (0.98, 0.67, and 0.57 at $\lambda=1$, 0.1, and 0). Here the reference is $\lambda=0$, because the correct-cue items have no uncued run. With $\lambda=0$ as the reference for both, the figures are 74\% and 76\%. A smaller Llama-3.1-8B-Instruct replication gives 85\% and 60\% (88\% and 60\% with $\lambda=0$ for both). It ran on 40 wrong-cue and 40 correct-cue items after a node failure stopped a first run. That first run had 55 wrong-cue items and no correct-cue arm, and it gives 86\% (89\% with $\lambda=0$ as the reference).

\subsection{Two-turn span masking}
\label{app:spanmask}
The scripted assistant reply after the user's challenge is a fixed turn that names another option in about 40\% of items. On items that the model answers correctly before the challenge, we set to zero the attention to both the challenge and this reply. This removes retraction (0 of 220 items in Llama-3.1-8B and 2 of 402 in Gemma-2-2b, against natural rates of 0.33 and 0.36), while masking the challenge alone leaves 0.30 and 0.27, and masking the reply alone leaves 0.36 and 0.18. With both spans masked, the effective context reduces to the question, the scripted correct first answer, and the final request, so the complete removal partly reflects this reduced context. Masking only in the six-layer band chosen in advance (layers 10 to 15 of 32 in Llama-3.1-8B-Instruct and 13 to 18 of 26 in Gemma-2-2b-it) removes retraction as completely as masking every layer (Llama 0.37 to 0.00, $n=451$, and Gemma 0.36 to 0.01, $n=402$), while random six-layer bands outside the circuit layers leave it at 0.36. The reader heads' attention mass on the two spans predicts retraction at AUROC 0.75 (Llama) and 0.72 (Gemma).

\subsection{Two-turn head patching}
\label{app:headpatch}
\begin{figure*}[t]
\centering
\includegraphics[width=\textwidth]{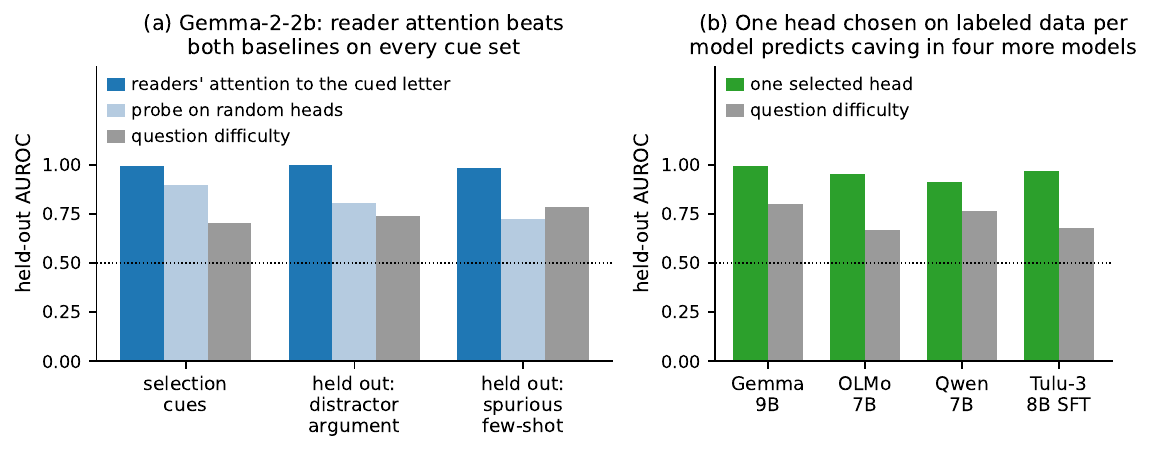}
\caption{\textbf{Attention to the cued letter separates caving from resisting items on held-out questions in five models.} \textbf{(a)}~Gemma-2-2b's fixed readers, on the cue types used for selection and on two held-out cue types. \textbf{(b)}~One head selected per model on a labeled training half. Only the OLMo-2 and Tulu-SFT heads are designated readers. The measure partly reads the answer that is forming (Appendix~\ref{app:attn}).}
\label{fig:monitor}
\end{figure*}

\begin{figure*}[t]
\centering
\includegraphics[width=0.85\textwidth]{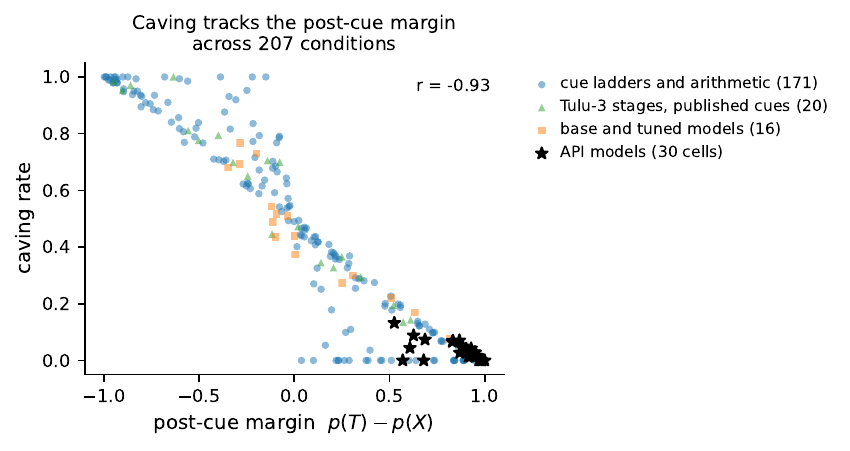}
\caption{\textbf{Across 207 open-weight conditions, caving falls close to one curve of the post-cue margin.} An indicator for base against tuned adds $\Delta R^2 = 0.004$, and the residual standard deviation is 0.12. The largest departures are uncued cells, where caving is zero by definition, and base-model arithmetic cells. Stars are the API models on the weak ladder, with one wording changed.}
\label{fig:marginlaw}
\end{figure*}

We patch the answer-position outputs of heads chosen in advance with their values from the single-turn prompt. These are three readers and the writer hub L17.H3 in Gemma, and three readers and two writers in Llama. The patch recovers 0.75 of naturally retracting items in Gemma-2-2b and 0.44 in Llama. Random head sets of the same size recover at most 0.034, and retention on items that do not retract is 0.99 to 1.00. The bar of 0.50 set in advance is met in Gemma and missed in Llama, where the writer heads alone recover 0.46 and the reader heads none. In Gemma the readers and writers each carry part of the effect (0.30 and 0.47).

\section{Additional results}
\label{app:additional}
\begin{figure}[t]
\centering
\includegraphics[width=\columnwidth]{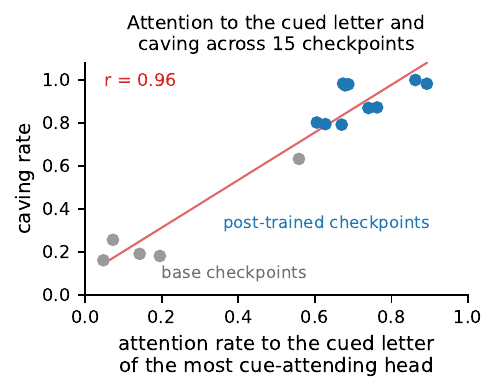}
\caption{\textbf{Attention to the cued letter tracks caving across 15 checkpoints, mostly through the difference between base and post-trained checkpoints.} The x-axis uses the most cue-attending head in a fixed layer band and partly reflects the chosen answer. The ten post-trained checkpoints alone give $r=0.58$ ($p=0.08$), and each family's final checkpoint caves on nearly all of these items by construction.}
\label{fig:fingerprint}
\end{figure}

The results in this section support no claim in the main text. We report them briefly, with their caveats.

\subsection{A reader-attention monitor}
\label{app:monitor}
In five models, attention to every occurrence of the cued letter separates caving from resisting items on held-out questions. Held-out AUROC is 0.91 to 0.995 on the cue types used for selection, and 0.985 and 0.998 on two held-out cue types in Gemma-2-2b (Figure~\ref{fig:monitor}). Gemma-2-2b uses its fixed reader heads, and Gemma-2-9b, Qwen-2.5-7B, OLMo-2, and Tulu-3-SFT use one head selected on a labeled training half. In Llama-3.1-8B, two fixed heads found by patching reach 0.869, about the same as a probe on the outputs of random heads (0.868). The measure counts the cued letter anywhere in the prompt and is read in the same forward pass whose logits define the caving label, so part of the signal is the answer that is already forming (Appendix~\ref{app:attn}). The tests on held-out cue types also reuse questions, and the fixed Gemma-2-2b and Llama heads were found by patching on flips inside the evaluation pool.

\subsection{The margin relationship}
\label{app:api}
Each condition pairs one checkpoint with one cue setting, including no cue. Across 207 open-weight conditions, caving follows the post-cue margin $p(T)-p(X)$ (Figure~\ref{fig:marginlaw}). This is consistent with splitting flips into a truth margin and a sensitivity to manipulation \citep{demarez2026decomposing}. Caving and the margin share the same output probabilities, so the correlation itself says little. The informative result is that the conditions fall close to one curve. We also ran the weak ladder on 88 items through OpenRouter on five API models (openai/gpt-4o-mini, openai/gpt-4o-2024-11-20, x-ai/grok-4.3, qwen/qwen3-235b-a22b-2507, and meta-llama/llama-4-maverick, after a smoke test fixed in advance dropped two more), scoring letters from the top 20 log-probabilities with letter forms summed and not renormalized. The probability of $T$ is missing for at most 1\% of items per cell, and that of $X$ for up to 72\% (Grok-4.3 without a cue). Their cells fall within the open-weight residual band (median absolute residual 0.105, inside the 0.15 set in advance), but each cell rests on at most nine caved items, and their pooled correlation ($-0.58$) misses the criterion of $r \le -0.8$ set in advance. Computed separately for each API model, $r \le -0.8$ holds in four of five models when the uncued cell is included, and in three of five without it. An API shows only behavior, so these results extend the relationship to API models and leave the mechanism claims untested there.

\subsection{A checkpoint fingerprint}
\label{app:fingerprint}
For each of 15 checkpoints of five models, we take the head in a fixed layer band that attends most to the cue. Its attention rate to the cued letter tracks caving at $r=0.96$ (Figure~\ref{fig:fingerprint}), mostly through the difference between base and tuned models. If we choose the head once on the pretrained base and keep it across checkpoints, the correlation is $r=-0.47$ (10 checkpoints). We also fit a model in which susceptibility is one strength per stage times one strength per cue. It reconstructs left-out cells of the stage-by-cue susceptibility grid at RMSE 0.080 (Tulu) and 0.041 (OLMo), compared with data standard deviations of 0.276 and 0.200. Most of the structure of the grid, however, is the profile of cue strengths, and the grids are small (4$\times$4 and 3$\times$4). A fit on half of the questions predicts each held-out item's cue-induced margin shift from its reader attention rate at $r=0.84$ (Tulu) and $0.85$ (OLMo), compared with $r=0.08$ for the uncued margin (Tulu). Both the checkpoint-level and the item-level attention measures partly track the answer (Appendix~\ref{app:attn}).

\section{Rules set in advance}
\label{app:prereg}
We recorded all rules with self-reported dates in a project log that has no external timestamps. The weight-copy tests had no numerical threshold.

\subsection{Stage tests}
Of the 13 stage tests, the 32B published-cue test, the 70B test, and the arithmetic tests had decision rules fixed before their data were collected. The other stage tests were not specified in advance. The graded tests use a weak-cue set that we built after the strong-cue ladder saturated, a ladder that we had specified in advance. In that ladder, caving was 0.98 to 1.00 in both stages at two of its three strongest wordings, and the third wording (0.90 compared with 0.93) gave $t=1.4$. The reported values apply four analysis choices made afterward, and none of them changes the outcome of a stage test. First, we exclude cue-less items (32B, and the published-cue attention test, whose $t$ as originally run was 0.96 in Tulu and 2.62 in OLMo). Second, we use the cued-minus-uncued susceptibility contrast (70B and arithmetic). Third, for arithmetic we average the four mid-strength wordings and do not use the wordings that each model's headroom filter selected. Fourth, we average repeated arithmetic problems into one unit. The rules also stated that a 70B null would leave the scale claim at 32B (the 70B test gave the one null), that the reasoning comparison would be a binary contrast unless items of intermediate derivability fell strictly between the arithmetic and knowledge rates of the original run (they did not), and how candor rates in the additional recipe would be read (the cell proved too small to quantify).

\subsection{Follow-up experiments}
Rules were fixed before the full runs of attention matching, the layer-set sweep, the reader-head rerun, both role-control runs, the leave-one-cue-out test, the published-cue attention test, the attention-pattern swap and its extensions, the query-key-value split, the fixed query shift, the pretrained-base band test, the evaluation of the fine-tuning runs on the unselected set, the item-matched chain-of-thought comparison, the two-turn tests, and the item-disjoint rerun of the activation directions. Transfer counts as substantial at $M \ge 0.6$, partial at $0.3$ to $0.6$, and null below $0.3$ or with a CI spanning zero. After a generator error in the first swap run, the rules also fixed the seeds and the exclusion rule for the clean random sets. The rule for the item-disjoint rerun was that Figure~\ref{fig:switch} would show its numbers if the effect survived. It survived. The Gemma weight copy, the second-family role control, and the original two-turn experiment had only an expected pattern recorded in advance. The Gemma weight copy met it, and the second-family role control showed it more weakly.

\subsection{Rules that were not met}
We report the thirteen rules that were not met, each with its result.
\par\noindent\hangindent=1.2em\hangafter=1 $\bullet$~Strong-cue ladder, a change in caving with $|t|>2$ at its three strongest wordings (two were at 0.98 to 1.00 in both stages, and the third gave $t=1.4$).
\par\noindent\hangindent=1.2em\hangafter=1 $\bullet$~Reasoning comparison, chain-of-thought caving on items of intermediate derivability between the DPO arithmetic and knowledge rates of the original run (0.04 and 0.12, outside the band of 0.18 to 0.68).
\par\noindent\hangindent=1.2em\hangafter=1 $\bullet$~Attention matching, all four cells reach 0.6 (Tulu downward is partial, 0.51, Appendix~\ref{app:gainmatch}).
\par\noindent\hangindent=1.2em\hangafter=1 $\bullet$~Reader-head rerun, substantial transfer, with transfer by random heads below half the transfer by reader heads (OLMo partial, 0.36 and 0.37, and Tulu null, Appendix~\ref{app:gainlocal}).
\par\noindent\hangindent=1.2em\hangafter=1 $\bullet$~Layer-set sweep, a layer set carries the difference if its $M$ reaches 0.6 of the network-wide $M$, its CI excludes zero, and its matched control sets stay below half of its own $M$ (OLMo downward reaches 0.54 of the network-wide $M$, Appendix~\ref{app:gainlocal}).
\par\noindent\hangindent=1.2em\hangafter=1 $\bullet$~Pretrained-base band test, a middle-band response that rises at every step and is larger than the responses of the low and high bands (in the OLMo base it goes from 0.34 to 0.32 between $\lambda=2$ and 4, Appendix~\ref{app:baseband}).
\par\noindent\hangindent=1.2em\hangafter=1 $\bullet$~First swap wave, the middle band transfers more than twice the largest control set without middle-band or reader layers (not met in Tulu, Appendix~\ref{app:swap}).
\par\noindent\hangindent=1.2em\hangafter=1 $\bullet$~First held-out-cue run, both controls per cell stay below 0.3 (one Tulu control reached 0.47, Appendix~\ref{app:swap}).
\par\noindent\hangindent=1.2em\hangafter=1 $\bullet$~Fixed query shift, a threshold of 0.3 (OLMo 0.23 and 0.24, Appendix~\ref{app:qkv}).
\par\noindent\hangindent=1.2em\hangafter=1 $\bullet$~Query-key-value split, an expected null on the value side (Tulu 0.38 and 0.33, Appendix~\ref{app:qkv}).
\par\noindent\hangindent=1.2em\hangafter=1 $\bullet$~Published-cue attention test, a positive change with $|t|>2$ (Tulu $t=1.1$, Figure~\ref{fig:rlstep}b, Appendix~\ref{app:rank}).
\par\noindent\hangindent=1.2em\hangafter=1 $\bullet$~Two-turn head patching, a bar of 0.50 (Llama 0.44, Appendix~\ref{app:headpatch}).
\par\noindent\hangindent=1.2em\hangafter=1 $\bullet$~API models, a pooled correlation of $r \le -0.8$ (pooled $-0.58$, Appendix~\ref{app:api}).
\par

\section{Additional related work}
\label{app:related}
\subsection{Mechanistic studies of post-training}
Related studies compare RL and SFT features trained from one base on identical data \citep{shi2026rlfeatures}, and find instruction representations acting as circuit selectors across OLMo-2 base, SFT, and DPO checkpoints \citep{bigoulaeva2026patches}. Others report higher activation intensity and less concentrated edge activations after online RL (PPO, GRPO), with weaker or inconsistent changes after DPO \citep{zhang2026rlcircuitry}. Further studies report that gains in context grounding from SFT and DPO largely reuse machinery present in the starting model \citep{gupta2026grounding}, that fine-tuning strengthens an existing entity-tracking circuit \citep{prakash2024finetuning}, that it can act as a thin wrapper around existing abilities \citep{jain2024mechanistically}, and that DPO routes around a pretrained ability and leaves it in place \citep{lee2024mechanistic}. Others follow how circuits change across fine-tuning checkpoints \citep{wang2025finetuning,chen2026oldmaps}.

\subsection{Locating sycophancy and cue-following}
Sycophancy has been located in heads found by path patching and then fine-tuned \citep{he2024pinpoint}, in head activations that can be decoded linearly \citep{genadi2026sycophancy}, and in steering directions that separate sycophantic behaviors \citep{vennemeyer2025distinct}. It has also been located in error-detection heads present in a base model and kept across later aligned releases \citep{pandey2026sharedcircuit}, in modes that are distinct in representation but share most of their attention heads \citep{gotta2026modes}, and in late layers where user opinions override factual knowledge \citep{wang2026truth}. Each of these studies uses released checkpoints or pairs of models before and after a single alignment or fine-tuning step. None follows the successive stages of one post-training pipeline. \citet{kumarappan2026notjustrlhf} find that base models yield to simulated peer pressure as much as or more than their instruct versions in 10 of 12 cells of family and condition, and they locate the effect in Llama-3.1-8B-Instruct in attention in layers 14 to 18. Tracing persona directions finds sycophancy directions forming early in pretraining and, across the SFT, DPO, and RLVR checkpoints of OLMo-3, persona suppression concentrated at the DPO stage \citep{moskvoretskii2026persona}. Related heads copy wrong labels from demonstrations \citep{halawi2024overthinking} and select multiple-choice answer letters \citep{lieberum2023circuit,wiegreffe2025answer}, and the heads we study may overlap with such answer-selection heads. Different biases can also occupy different directions in the hidden states \citep{gupta2026directions}. Our comparison with the format held fixed suggests one reason why chat-format comparisons of base and final models, including direction-level work that attributes susceptibility to alignment tuning \citep{gupta2026directions}, can seem to show that alignment builds the circuit. Consistency training partly transfers to held-out biases \citep{chua2024bct,imran2026ratematching} and, in one model, from jailbreaks to sycophancy \citep{gautam2026stack}, which a shared circuit could explain.

\subsection{Interventions on attention}
Rescaling attention weights or head outputs can shift models between facts in the prompt and facts in memory \citep{yu2023characterizing,ortu2024competition}, and attention steering can help them follow highlighted text \citep{zhang2023pasta}. Concurrent work uses attention steering to reduce sycophancy \citep{he2026spae}. Suppressing unsupported agreement tends to harm updating on real evidence \citep{ma2026suppression}, and resisting wrong suggestions and accepting correct ones are distinct abilities \citep{sinha2026sycobench}. \citet{he2026spae} report that suppressing attention to user pressure while also amplifying attention to supplied context lowers pressure following and improves updating on that context, although their method suppresses a user's stated answer whether or not it is correct. We measure this trade-off for scaling attention to the cue (\S\ref{sec:dial}).

\end{document}